\documentclass{article}

\usepackage[preprint]{neurips_2026}

\usepackage[utf8]{inputenc}
\usepackage[T1]{fontenc}
\usepackage{hyperref}
\usepackage{url}
\usepackage{booktabs}
\usepackage{amsfonts}
\usepackage{nicefrac}
\usepackage{microtype}
\usepackage{xcolor}

\usepackage{graphicx}
\usepackage{subcaption}
\usepackage{amsmath}
\usepackage{amssymb}
\usepackage{mathtools}
\usepackage{amsthm}
\usepackage{tikz}
\usetikzlibrary{positioning, arrows.meta, calc, fit, backgrounds, shapes.geometric, decorations.pathreplacing}

\usepackage{algorithm}
\usepackage{algorithmic}

\usepackage[capitalize,noabbrev]{cleveref}

\theoremstyle{plain}

\theoremstyle{definition}

\theoremstyle{remark}

\usepackage[disable,textsize=tiny]{todonotes}

\title{Physics-Informed Tracking (PIT)}

\author{%
  Emil Hovad \\
  Department of Applied Mathematics \\
  and Computer Science (DTU Compute) \\
  Technical University of Denmark \\
  \texttt{emilh@dtu.dk} \\
  \And
  Allan Peter Engsig-Karup \\
  Department of Applied Mathematics \\
  and Computer Science (DTU Compute) \\
  Technical University of Denmark \\
  \texttt{apek@dtu.dk} \\
}

\begin{document}

\maketitle

\begin{abstract}

We propose Physics-Informed Tracking (PIT),
a video-based framework for tracking a single particle from video,
where a neural network autoencoder localizes a particle
as a heatmap peak (landmark) and a differentiable physics module embedded in the autoencoder
constrains several landmarks over time (a trajectory)
to satisfy known dynamics. The novel Physics-Informed Landmark Loss (PILL)
compares this predicted trajectory back against the landmarks,
enforcing physical consistency without labels.
Its supervised variant (PILLS) instead compares the prediction
against ground-truth position, velocity, and bounce from simulation,
enabling end-to-end backpropagation.

To support supervised and unsupervised learning,
we use an autoencoder with a split bottleneck that separates
A) tracking-related structure via landmark heatmaps from
B) background noise and subsequent image reconstruction.
We evaluate a replicated $2^6$ factorial design ($n=4$ replicates, 64 configurations),
showing that PILLS consistently achieves
sub-pixel tracking accuracy for the bilinear and physics-refined decoder outputs under both clean and noisy conditions.

\end{abstract}

\section{Introduction}
%Deep learning has revolutionized machine learning by enabling models to automatically learn complex patterns from data, typically with the back-propagation algorithm \cite{rumelhart1986a}. 
%This has lead to significant advancements across domains like computer vision, natural language processing, and speech recognition \cite{ Goodfellow-et-al-2016}.
%The resurgence of deep learning in the early 2000s was fueled by increased computational power, availability of large datasets, and the development of innovative neural network architectures, with 
%CNNs, initially introduced by \cite{Fukushima1980NeocognitronAS}, were inspired by the human visual cortex and excel in image processing by capturing local spatial relationships. Early CNNs, such as LeNet for handwritten digit recognition \cite{le1989handwritten}, were limited by computational resources. However, the advent of powerful GPUs and large-scale datasets like ImageNet \cite{deng2009imagenet} spurred a resurgence in CNN research. 
%Convolutional Neural Networks (CNNs) has been a cornerstone in computer vision. initially introduced by \cite{Fukushima1980NeocognitronAS}, were inspired by the human visual cortex and further developed by LeNet for handwritten digit recognition \cite{le1989handwritten}. 

% First block - general background on deep learning and CNNs, with a focus on their application to computer vision and object tracking.
Tracking objects in video is a core problem in computer vision. 
Among the first successful and efficient deep learning approaches to object detection were Faster R-CNN \cite{renFasterRCNNRealTime2015} and YOLO \cite{redmonYouOnlyLook2016}, 
while architectures with skip connections, notably ResNet \cite{heDeepResidualLearning2016} and U-Net \cite{ronnebergerUNetConvolutionalNetworks2015}, 
have become standard for dense prediction tasks. More recently, keypoint-based methods, also called landmark methods, 
that localize objects as peaks in heatmaps have gained attention. CenterNet \cite{duanCenterNetKeypointTriplets2019b} detects objects as keypoint triplets, while Zhou et al.\ \cite{zhouObjectsPoints2019} propose a simpler heatmap-based formulation where objects are represented as center points. Our work is inspired by the latter approach, using heatmap peaks as landmark positions. 
% -next block - background on autoencoders. 
Autoencoders learn compact latent representations without labels \cite{bengioDeepLearningRepresentations2012}, 
and denoising autoencoders (DAEs) improve robustness by reconstructing clean data from noisy inputs \cite{vincentExtractingComposingRobust2008}. 
Our work builds on the skip connection, landmark design and autoencoder principles from these architectures and extends them with physics-informed constraints for more precise landmark localization in particle tracking.

% Related work.
% -next block - background on autoencoders and physics-informed neural networks, with a focus on their applications to tracking and dynamics modeling.
\subsection{Related work in auto-encoder tracking and scientific machine learning}
In tracking, autoencoder-based methods have been used for state estimation \cite{xuWassersteinDistanceBasedAutoEncoder2021} and high-speed feature compression \cite{choiContextAwareDeepFeature2018}, 
however none of these incorporate physical constraints into the tracking process.
Physics-informed neural networks (PINNs) \cite{raissi2019physics}, 
developed for continuum mechanics and partial differential equations, embed physical laws directly into the learning process,
enabling data-driven solutions where labeled data is scarce. This principle has been extended to autoencoders: 
physics-informed autoencoders (PIAEs) enforce physical consistency in latent representations, 
for example through the Koopman operator which linearizes nonlinear dynamics \cite{riceAnalyzingKoopmanApproaches2021}.

The physics-informed trajectory autoencoder (PITA) \cite{fischerPITAPhysicsInformedTrajectory2024} 
is an autoencoder that takes vehicle trajectory coordinates as explicit input and integrates a kinematic bicycle model as a physical regularization to produce smooth, 
physically plausible reconstructions. Notably, Fischer et al. state that, to their knowledge, no prior autoencoder has incorporated physical constraints into trajectory encoding. In contrast to PITA, PIT must first extract particle coordinates implicitly from raw video frames via learned heatmaps before applying physics constraints. 
 \cite{erichsonPhysicsinformedAutoencodersLyapunovstable2019} introduces a physics-informed autoencoder for fluid flow prediction from visual snapshots, 
 where a skip connection separates the dynamics model from an identity-preserving component and a Lyapunov stability prior constrains the learned dynamics. 
PIT adopts a similar separation principle with its split bottleneck, but replaces the stability prior with explicit motion equations and operates 
on sparse landmark heatmaps rather than dense flow fields. 

Closely related, \cite{kienzleLearningMonocular3D2023} learn
monocular 3D object localization from 2D labels using the physical laws of motion:
their Position Estimation Network predicts 2D heatmaps and a depthmap from a single image,
and a Physics Aware Forecast module (a Neural ODE with soft potential walls)
supervises the depth via a future-frame consistency loss.
PIT differs fundamentally:
(i)~their problem is single-image 3D localization,
whereas PIT performs temporal landmark tracking in 2D;
(ii)~their heatmaps are trained with GT 2D labels,
whereas PIT's PILL is fully unsupervised;
(iii)~their physics module is discarded at inference (single-image test time),
whereas PIT's differentiable Velocity-Verlet module is active at both training and inference and produces position, velocity, and bounce outputs from a single forward pass.

SINDy \cite{bruntonDiscoveringGoverningEquations2016} discovers governing equations from data via sparse regression, and a comprehensive treatment of data-driven dynamical systems is given in \cite{bruntonDataDrivenScienceEngineering2022}.
While these methods enforce physical consistency in latent representations or trajectory predictions, none apply physics-informed constraints directly to visual landmark tracking with an autoencoder design.
PIT bridges this gap by introducing physics-informed losses (PILL and PILLS) that constrain landmark trajectories to satisfy known motion dynamics, enabling end-to-end learning of position, velocity, and bounce estimation.
Additionally PIT combines an autoencoder with a structured bottleneck that separates tracking landmarks from background noise, enabling both supervised and unsupervised learning.

\subsection{Contributions}

We propose \emph{Physics-Informed Tracking (PIT)}, whose contributions are as follows.
First, we introduce a \emph{split autoencoder bottleneck} that separates (A)~tracking-related landmark heatmaps, whose maxima correspond to particle locations, from (B)~a background/noise component used for image reconstruction. This design is inspired by the separation of dynamics and identity in \citet{erichsonPhysicsinformedAutoencodersLyapunovstable2019}, and we refer to the landmark outputs as \emph{Autoencoder Landmark Outputs} (AELO), or \emph{AELOS} when ground-truth supervision is applied.
Second, we introduce the \emph{Physics-Informed Landmark Loss} (PILL), an unsupervised loss that constrains landmark trajectories to satisfy known physical laws --- e.g.\ gravity-driven parabolic motion --- without requiring ground-truth labels. PILL is conceptually related to PINNs but applied to landmark tracking rather than field regression.
Third, we introduce a supervised variant, \emph{PILL Supervised} (PILLS), in which network-predicted landmarks are projected into a physical state space via a differentiable physics module that evolves the system dynamics; all operators are part of the computational graph, enabling end-to-end supervised learning of position, velocity, and bounce dynamics.
Finally, a key advantage of the physics-informed approach is that PILL and PILLS provide not only refined position estimates but also velocity predictions and bounce timing/position, all from a single forward pass of the differentiable physics module --- physical state predictions that are not available from standard heatmap-based tracking methods.

We evaluate on simulated ball trajectories under clean and noisy conditions using CenterNet-style \cite{duanCenterNetKeypointTriplets2019b} heatmap supervision as our baseline, a multi-scale decoder with skip connections for landmark refinement, and a replicated $2^6$ factorial design ($n=4$ replicates, 64 configurations). Results show that physics-informed landmark constraints consistently improve tracking performance over standard heatmap training.

% CODE: Data generation script: src/scripts/a_generate_data.py
% CODE: Ball trajectory simulation (simulate_ball_motion_analytical) in a_generate_data.py:100-169
% CODE: Video frame generation with noise in a_generate_data.py:172-200
% CODE: Dataset saving pipeline in a_generate_data.py:204-247
\section{Data}\label{Chap:data}
The dataset is comprised of synthetic video sequences
created by simulating the motion of a ball following
a parabolic trajectory with inelastic boundary
collisions (Table~\ref{tab:experiment_setup}).
%This simulation accounts for the effects of gravity, generating the ball's positions over time from a given initial position and predefined velocity vector. 

\subsection{The physics: gravity-only model for a ball}
Starting from Newton's second law
\begin{equation}
m {\bf a} = {\bf F}, \quad t>0,
\end{equation}
we assume a constant gravity-only force model, ${\bf F}$ ($i=0$), implying ${\bf a}=(a_x,a_y)^T=(0,g)^T$ where the gravitational acceleration is assumed to be constant (positive in this setup, as the \( y \)-axis increases downward). Here, \( t \) [s] represents time, which is to be discretized into frames.

The ball is launched from an initial position \(\mathbf{x_0} =(x_0, y_0)\) [m] with an initial velocity \(\mathbf{v_0} = (v_{x0}, v_{y0})\) [m/s]. The ball's motion is then governed by the following equations:
\begin{subequations}
\begin{equation}
x(t) = x_0 + v_{x0} \cdot t,
\end{equation}
\begin{equation}
y(t) = y_0 + v_{y0} \cdot t + \frac{g}{2} \cdot t^2,
\end{equation}
\end{subequations}
%Where \( g \) is the acceleration due to gravity (positive in this setup, as the \( y \)-axis increases downward), and \( t \) represents time, which is discretized into frames. 

Inelastic collisions with the image boundaries are modeled by velocity reflection. Upon collision with a vertical boundary, the horizontal velocity component is updated as
\begin{equation}
v_x^{t_1} = - e\, v_x^{t_0},
\end{equation}
and upon collision with a horizontal boundary, the vertical velocity component is updated as
\begin{equation}
v_y^{t_1} = - e\, v_y^{t_0},
\end{equation}
where $t_0$ and $t_1$ denote the pre- and post-collision velocities, respectively, and $e$ is the coefficient of restitution.

%The simulation stops when the ball reaches the ground, defined by the condition \( y(t) \geq H \), where \( H \) is the height of the simulation area. After the ball reaches this point, the simulation does not account for any post-bounce motion.
\subsection{Random Initial Position and Velocity Generation}
To introduce variability in the ball's dynamics during the synthetic video simulations, both the initial velocities and the initial positions are randomized. Each initial velocity $\mathbf{v_0} = (v_{x0}, v_{y0})$ consists of a horizontal component $v_{x0}$ and a vertical component $v_{y0}$, both drawn from predefined ranges, while the initial position $\mathbf{x_0} = (x_0, y_0)$ is also randomly chosen inside the simulation box.

%A fixed random seed (\( \text{seed} = 42 \)) is used to ensure the reproducibility of the experiments... Emil check in code!

\paragraph{\textbf{Random initial conditions.}}
For each sequence, the initial position and velocity are sampled independently using the python package NumPy \cite{harrisArrayProgrammingNumPy2020}. Initial positions are drawn uniformly inside the valid image region,

\[
x_0, y_0 \sim \mathcal{U}\!\big(r,\; W-r\big),
\]
\[
v_{x0}, v_{y0} \sim \mathcal{U}\!\big(-v_{\max},\; v_{\max}\big).
\]

This ensures that the ball center is initialized fully inside the image domain. All random sampling is performed using fixed pseudo-random seeds with a NumPy seed of $42$, to ensure reproducibility.
%\[
%(x_0, y_0) \sim \texttt{np.random.randint}(r,\; W-r),
%\]
%ensuring that the ball starts fully inside the frame. Initial velocities are sampled as
%\[
%(v_{x0}, v_{y0}) \sim \texttt{np.random.randint}(-10,\;10),
%\]
%yielding integer-valued horizontal and vertical velocity components. All random sampling uses fixed NumPy seed of $42$.% and PyTorch seeds for reproducibility.

\paragraph{Projection onto the image plane:}
The ball’s physical position [m], \((x_{\text{physical}}(t), y_{\text{physical}}(t))\), is projected onto the pixel grid. In a general case, the projection from physical space to pixel coordinates can be expressed as:

\[
x_{\text{pixel}}(t) = \frac{x_{\text{physical}}(t)}{S_x}, \quad y_{\text{pixel}}(t) = \frac{y_{\text{physical}}(t)}{S_y}
\]

where \(S_x\) and \(S_y\) are scaling factors representing the number of meters per pixel in the \(x\) and \(y\) directions, respectively.

\subsection{Experimental Setup, Data Splitting and Video Generation}
Table~\ref{tab:experiment_setup} summarizes all simulation and data parameters. The synthetic video data are split into training (100 sequences), validation (50 sequences), and test (100 sequences) sets. All splits share the same physics and imaging parameters but differ in the random initial conditions, which are sampled from continuous uniform distributions over positions and velocities. No sequence appears in more than one split. The training set is used for model learning, the validation set for model selection (best epoch per metric), and the test set for final evaluation only.

\begin{table}[h!]
\centering
\small
\caption{Experimental setup for training, validation and test data.}
\label{tab:experiment_setup}
\begin{tabular*}{\columnwidth}{@{\extracolsep{\fill}}lccc@{}}
\toprule
\textbf{Parameter} & \textbf{Train} & \textbf{Val} & \textbf{Test} \\
\midrule
Image size $H\!=\!W$ [px]           & $224$       & $224$       & $224$       \\
Scale $S_x\!=\!S_y$ [m/px]          & $0.02$      & $0.02$      & $0.02$      \\
Physical domain $L_x\!=\!L_y$ [m]   & $4.48$      & $4.48$      & $4.48$      \\
Time step $\Delta t$ [s]            & $0.04$      & $0.04$      & $0.04$      \\
Ball radius $r$ [px] ($0.04$\,m)    & $2$         & $2$         & $2$         \\
Gravity $g$ [m/s$^2$]               & $9.81$      & $9.81$      & $9.81$      \\
Restitution $e$                     & $0.75$      & $0.75$      & $0.75$      \\
Max velocity $v_{\max}$ [m/s]       & $11.1$      & $11.1$      & $11.1$      \\
Frames per video ($1.6$\,s)         & $40$        & $40$        & $40$        \\
Noise levels $\sigma$               & $\{0,1\}$   & $\{0,1\}$   & $\{0,1\}$   \\
Number of sequences                 & $100$       & $50$        & $100$       \\
\bottomrule
\end{tabular*}
\end{table}

\paragraph{Video generation.}
The simulation uses a standard image coordinate system, where the origin \((0, 0)\) is at the top-left corner of the frame. The \(x\)-axis increases from left to right, and the \(y\)-axis increases from top to bottom. The ball's motion is mapped accordingly over time to these coordinates.

For each ball position \(\mathbf{p_i}\), a corresponding frame \(F_i\) is generated by first initializing a blank canvas of dimensions \(H \times W\), followed by the addition of static Gaussian noise. The filled circle representing the ball is drawn at position \(\mathbf{p_i}\), with a radius \(r\), superimposed on the noisy background. The static noise is generated once and remains consistent across all frames, following the distribution:
\[
\epsilon = \mathcal{N}(0, \sigma^2),
\]
where \(\sigma\) denotes the standard deviation of the noise and the resulting noise image is denoted by \(\epsilon\) which is added to each frame to simulate a consistent noisy background. The generated frames are stacked into a 3D array that represents the entire video sequence.
%This sequence is encoded using the MP4 codec, with a frame rate of 1 frame per second, with a time step of $\Delta t = 1$.

\begin{figure}[ht]
\centering
\includegraphics[width=0.9\columnwidth]{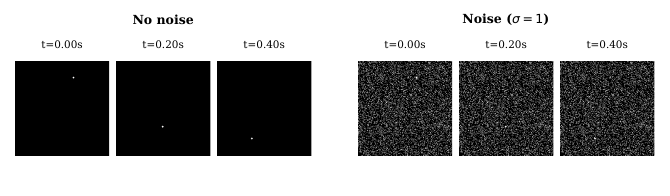}
\caption{Example input frames at frames 0, 5, and 10 (corresponding to $t = 0$, $0.20$, and $0.40$\,s with $\Delta t = 0.04$\,s).
Left three panels: clean frames ($\sigma=0$). Right three panels: noisy frames ($\sigma=1$).
The ball (radius $r=2$ pixels) is clearly visible without noise but nearly
indistinguishable from the background with noise.}
\label{fig:data_examples}
\end{figure}

Examples of video frames with varying noise levels are presented in Figure~\ref{fig:data_examples}.

\section{Method}\label{Theory}
In this section, we describe the PIT
encoder-decoder architecture, the training
procedure with six binary factors (A--F),
and the evaluation protocol.

\subsection{PIT: Encoder-Decoder Architecture}

\begin{figure*}[t]
\centering
\resizebox{0.92\textwidth}{!}{%
\begin{tikzpicture}[
    >=latex,
    every node/.style={font=\sffamily},
    block/.style={draw, rounded corners=2pt, minimum height=0.45cm, minimum width=1.1cm, font=\sffamily\scriptsize, align=center, thick},
    input/.style={block, fill=gray!8},
    encoder/.style={block, fill=blue!6},
    head/.style={block, fill=blue!10},
    fusion/.style={block, fill=blue!6},
    physics/.style={block, fill=red!8},
    decoder/.style={block, fill=green!8},
    heatmap/.style={block, fill=orange!10},
    outp/.style={block, fill=gray!5, minimum width=1.8cm},
    loss/.style={draw, dashed, rounded corners=2pt, fill=white, minimum height=0.35cm, font=\sffamily\scriptsize, align=center, thin, minimum width=1.6cm},
    arr/.style={->, thick, black!80},
    skiparr/.style={->, densely dashed, thick, black!50},
    lossarr/.style={->, thin, densely dashed, black!40},
    annot/.style={font=\sffamily\tiny, text=black!60},
]

% ============================================================
% INPUT FRAMES (y = 0)
% ============================================================
\node[input] at (-3.5, 0) (f1) {$I_{t-1}$};
\node[input] at (-2.0, 0) (f2) {$I_{t}$};
\node[input] at (-0.5, 0) (f3) {$I_{t+1}$};
\node[annot, above=3pt of f2] {$1{\times}224{\times}224$};
\node[annot, font=\sffamily\scriptsize\bfseries, above=12pt of f2] {Image inputs};

% ============================================================
% SHARED ENCODER (y = -1.0)
% ============================================================
\node[encoder, minimum width=3.2cm] at (-2.0, -1.0) (enc) {Shared Encoder (Conv--ResBlock--Pool $\times 2$)};

\draw[arr] (f1.south) -- ++(0,-0.12) -| ([xshift=-0.35cm]enc.north);
\draw[arr] (f2.south) -- (enc.north);
\draw[arr] (f3.south) -- ++(0,-0.12) -| ([xshift=0.35cm]enc.north);

% ============================================================
% SPLIT POINT (y = -1.4)
% ============================================================
\node[fill=black, circle, inner sep=1.2pt] at (-2.0, -1.4) (split) {};
\draw[arr] (enc.south) -- (split);

% ============================================================
% HEADS (y = -2.05)
% ============================================================
\node[head] at (-3.5, -2.05) (ch) {Center Head\\$1{\times}1$ conv};
\node[head] at (-0.5, -2.05) (rh) {Residual Head\\$3{\times}3$ conv};

\draw[arr] (split) -| (ch.north);
\draw[arr] (split) -| (rh.north);

\node[annot] at (-4.0, -2.55) {$\downarrow H^{56}$};
\node[annot] at (0.0, -2.55) {$\downarrow R^{56}$};

% ============================================================
% LATENT FUSION (y = -3.2)
% ============================================================
\node[fusion, minimum width=2.2cm] at (-2.0, -3.2) (fuse) {Latent Fusion ($1{\times}1$ conv)};

\draw[arr] (ch.south) -- ++(0,-0.25) -| ([xshift=-0.25cm]fuse.north);
\draw[arr] (rh.south) -- ++(0,-0.25) -| ([xshift=0.25cm]fuse.north);

% ============================================================
% SCALE 56 (y = -4.2)
% ============================================================
\node[heatmap] at (1.2, -4.2) (h56) {$H^{56}$};
\node[block, fill=orange!5] at (3.2, -3.9) (b56) {coarse-to-fine};
\node[block, fill=orange!5] at (3.2, -4.5) (harg56) {hard-argmax};
\node[physics] at (5.3, -3.9) (p56) {Physics ($g$,$e$)};
\node[outp]    at (7.5, -3.9) (o56) {$\hat{\mathbf{p}}^{56}\!,\hat{\mathbf{v}}^{56}\!,\hat{b}^{56}$};

\draw[arr] (fuse.south) -- ++(0,-0.25) -| node[pos=0.75, right, annot] {$\sigma$} (h56.north);
\draw[arr] (h56.east) -- ++(0.3,0) |- (b56.west);
\draw[arr] (h56.east) -- ++(0.3,0) |- (harg56.west);
\draw[arr] (b56.east) -- node[above, annot] {$\hat{\mathbf{c}}^{56}_{\mathrm{bil}}$} (p56.west);
\draw[arr] (p56.east) -- (o56.west);
\node[annot, anchor=west] at (4.0, -4.5) {(eval only)};

% Losses — all aligned at x=9.8
\node[loss] at (9.8, -3.9) (l56) {$\mathcal{L}_{\text{PILL}}^{56}$, $\mathcal{L}_{\text{PILLS}}^{56}$};
\draw[lossarr] (o56.east) -- (l56.west);

\node[loss] at (9.8, -4.8) (rg56) {$\mathcal{L}_{\text{hm}}^{56}$};
\draw[lossarr] (h56.south) -- ++(0,-0.45) -| (9.0, -4.8) -- (rg56.west);

\node[annot] at (-5.0, -4.2) {$56^2$};

% ============================================================
% DECODER 56 -> 112 (y = -5.4)
% ============================================================
\node[decoder, minimum width=2.6cm] at (-2.0, -5.4) (d112) {Decoder $56 \!\to\! 112$ (Conv--Up--ReLU)};

\draw[arr] (fuse.south) -- ++(0,-0.25) -- (-2.0, -5.05) -- (d112.north);

\node[heatmap, font=\sffamily\tiny] at (-1.4, -4.2) (h56skip112) {$\uparrow\!H^{56}$};
\draw[skiparr] (h56.west) -- (h56skip112.east);
\draw[skiparr] (h56skip112.south) -- ([xshift=0.6cm]d112.north);

% ============================================================
% SCALE 112 (y = -6.4)
% ============================================================
\node[heatmap] at (1.2, -6.4) (h112) {$H^{112}$};
\node[block, fill=orange!5] at (3.2, -6.1) (b112) {biquadratic};
\node[block, fill=orange!5] at (3.2, -6.7) (harg112) {hard-argmax};
\node[physics] at (5.3, -6.1) (p112) {Physics ($g$,$e$)};
\node[outp]    at (7.5, -6.1) (o112) {$\hat{\mathbf{p}}^{112}\!,\hat{\mathbf{v}}^{112}\!,\hat{b}^{112}$};

\draw[arr] (d112.south) -- ++(0,-0.2) -| node[pos=0.75, right, annot] {$\sigma$} (h112.north);
\draw[arr] (h112.east) -- ++(0.3,0) |- (b112.west);
\draw[arr] (h112.east) -- ++(0.3,0) |- (harg112.west);
\draw[arr] (b112.east) -- node[above, annot] {$\hat{\mathbf{c}}^{112}_{\mathrm{bil}}$} (p112.west);
\draw[arr] (p112.east) -- (o112.west);
\node[annot, anchor=west] at (4.0, -6.7) {(eval only)};

\node[loss] at (9.8, -6.1) (l112) {$\mathcal{L}_{\text{PILL}}^{112}$, $\mathcal{L}_{\text{PILLS}}^{112}$};
\draw[lossarr] (o112.east) -- (l112.west);

\node[loss] at (9.8, -7.0) (rg112) {$\mathcal{L}_{\text{hm}}^{112}$};
\draw[lossarr] (h112.south) -- ++(0,-0.45) -| (9.0, -7.0) -- (rg112.west);

\node[annot] at (-5.0, -6.4) {$112^2$};

% ============================================================
% DECODER 112 -> 224 (y = -7.6)
% ============================================================
\node[decoder, minimum width=2.6cm] at (-2.0, -7.6) (d224) {Decoder $112 \!\to\! 224$ (Conv--Up--ReLU)};

\draw[arr] (d112.south) -- ++(0,-0.2) -- (-2.0, -7.25) -- (d224.north);

% Skip connections into decoder 224
\node[heatmap, font=\sffamily\tiny] at (-1.4, -6.4) (h56skip) {$\uparrow\!\uparrow\!H^{56}$};
\node[heatmap, font=\sffamily\tiny] at (0.0, -6.4) (h112skip) {$\uparrow\!H^{112}$};
\draw[skiparr] (h112skip.south) -- ++(0,-0.1) -| ([xshift=0.85cm]d224.north);
\draw[skiparr] (h56skip.south) -- ++(0,-0.1) -| ([xshift=0.45cm]d224.north);

% ============================================================
% SCALE 224 (y = -8.6)
% ============================================================
\node[heatmap] at (1.2, -8.6) (h224) {$H^{224}$};
\node[block, fill=orange!5] at (3.2, -8.3) (b224) {bicubic};
\node[block, fill=orange!5] at (3.2, -8.9) (harg224) {hard-argmax};
\node[physics] at (5.3, -8.3) (p224) {Physics ($g$,$e$)};
\node[outp]    at (7.5, -8.3) (o224) {$\hat{\mathbf{p}}^{224}\!,\hat{\mathbf{v}}^{224}\!,\hat{b}^{224}$};

\draw[arr] (d224.south) -- ++(0,-0.2) -| node[pos=0.75, right, annot] {$\sigma$} (h224.north);
\draw[arr] (h224.east) -- ++(0.3,0) |- (b224.west);
\draw[arr] (h224.east) -- ++(0.3,0) |- (harg224.west);
\draw[arr] (b224.east) -- node[above, annot] {$\hat{\mathbf{c}}^{224}_{\mathrm{bil}}$} (p224.west);
\draw[arr] (p224.east) -- (o224.west);
\node[annot, anchor=west] at (4.0, -8.9) {(eval only)};

\node[loss] at (9.8, -8.3) (l224) {$\mathcal{L}_{\text{PILL}}^{224}$, $\mathcal{L}_{\text{PILLS}}^{224}$};
\draw[lossarr] (o224.east) -- (l224.west);

\node[loss] at (9.8, -9.2) (rg224) {$\mathcal{L}_{\text{hm}}^{224}$};
\draw[lossarr] (h224.south) -- ++(0,-0.45) -| (9.0, -9.2) -- (rg224.west);

\node[annot] at (-5.0, -8.6) {$224^2$};

% ============================================================
% RECONSTRUCTION (y = -9.7)
% ============================================================
\node[outp] at (-2.0, -9.7) (rec) {$\hat{I}$ \;$(3{\times}224^2)$};
\node[annot, font=\sffamily\scriptsize\bfseries, below=5pt of rec] {Image outputs};
\draw[arr] (d224.south) -- (rec.north);

\node[loss] at (9.8, -9.7) (lae) {$\mathcal{L}_{\text{AE}}$, $\mathcal{L}_{\text{cone}}$};
\draw[lossarr] (rec.east) -- (lae.west);

% ============================================================
% BACKGROUND REGIONS: Encoder / Decoder
% ============================================================
\begin{scope}[on background layer]
    % Encoder region
    \fill[blue!3, rounded corners=4pt] (-4.8, 0.35) rectangle (0.8, -3.45);
    % Decoder region
    \fill[green!6, rounded corners=4pt] (-4.8, -5.15) rectangle (0.8, -9.95);
\end{scope}
\node[font=\sffamily\tiny\bfseries, text=blue!40, rotate=90] at (-4.95, -1.5) {ENCODER};
\node[font=\sffamily\tiny\bfseries, text=green!40!black, rotate=90] at (-4.95, -7.5) {DECODER};

\end{tikzpicture}
}% end resizebox
\caption{PIT network architecture. A shared encoder produces heatmaps $H$ and residual features $R$ at $56^2$. At each scale, differentiable expectation operators extract sub-pixel positions $B$ (coarse-to-fine at $56$, biquadratic at $112$, bicubic at $224$), which are fed to the physics module to produce refined positions $\hat{\mathbf{p}}$, velocities $\hat{\mathbf{v}}$, and bounce predictions $\hat{b}$. Hard argmax provides integer-pixel positions (eval only). Skip connections ($\uparrow\!H^{56}$, $\uparrow\!H^{112}$) refine decoder heatmaps. All losses are shown on the right.}
\label{fig:network_architecture}
\end{figure*}
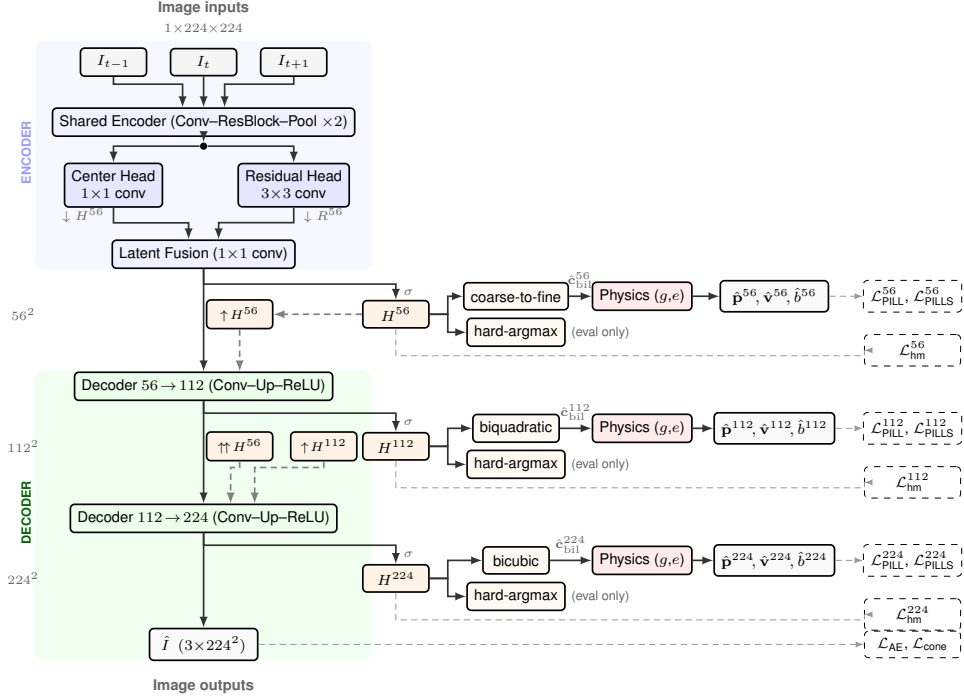

The full architecture is illustrated in
Figure~\ref{fig:network_architecture},
with layer specifications in
Table~\ref{tab:pit-architecture}
(Appendix~\ref{PIT_network}).
PIT processes video using a sliding window
of $T\!=\!3$ consecutive frames
$(t\!-\!1, t, t\!+\!1)$, each processed
independently --- no autoregressive rollout,
so errors do not accumulate across
the full sequence.

% PIT Network Architecture — Academic style (ICML/NeurIPS)
% Network on the left, processing chain flowing right, all losses aligned right

A shared encoder downsamples each frame
from $224^2$ to $56^2$ and splits
the latent into a tracking heatmap $H_t$
and a residual $R_t$ (background/noise),
fused via a $1\!\times\!1$ convolution.
Sub-pixel positions
$\hat{\mathbf{p}}_t \in \mathbb{R}^2$
are extracted by a differentiable
expectation operator and passed to a
Velocity-Verlet physics module with
bounce handling (\texttt{torch.where},
Appendix~\ref{fxy}), which outputs
physics-refined positions, velocities,
and bounce indicators.
A multi-scale decoder refines heatmaps
at $112$ and $224$ via skip connections,
applying the physics module at each
scale and reconstructing the input via
BCE loss.
Full layer specifications are in
Table~\ref{tab:pit-architecture}
(Appendix~\ref{PIT_network}).

\subsection{Training Procedure and Components}

Algorithm~\ref{alg:pit} summarizes the
training procedure for one window of
three frames.
We evaluate all $2^6\!=\!64$ combinations
of six binary factors using a replicated
factorial design ($n=4$ replicates).
Each factor controls a specific loss
or architectural component
(full derivations in
Appendix~\ref{app:loss_details}):

\noindent\textbf{A} (AELO/AELOS):
Passes landmark heatmaps into the
decoder via skip connections for
improved reconstruction.
If A=1, reconstruction gradients
($\mathcal{L}_{\text{AE}}$,
$\mathcal{L}_{\text{cone}}$)
flow back to the encoder through
the heatmap channels; if A=0 the
heatmaps are detached.

\noindent\textbf{B} (Noise bottleneck):
Passes unstructured encoder channels
(non-tracking) into the decoder for
background/noise reconstruction.
If B=1, reconstruction gradients
($\mathcal{L}_{\text{AE}}$, always active)
flow back to the encoder through the
residual channels; if B=0 the residual
is detached.

\noindent\textbf{C} (Heatmap supervision):
Enables focal landmark loss on Gaussian
target heatmaps.
C-only = CenterNet baseline.
Loss: $\mathcal{L}_{\text{hm}}$
(Eq.~\ref{eq:loss_hm}).

\noindent\textbf{D} (PILL):
Enables unsupervised physics loss ---
compares physics prediction against
heatmap landmarks.
Loss: $\mathcal{L}_{\text{PILL}}$
(Eq.~\ref{eq:loss_pill}).

\noindent\textbf{E} (PILLS):
Enables supervised physics loss ---
compares against ground-truth
position, velocity, and bounce
from simulation.
Loss: $\mathcal{L}_{\text{PILLS}}$
(Eq.~\ref{eq:loss_pills}).

\noindent\textbf{F} (Noise condition):
$\sigma\!=\!0$ (clean) or
$\sigma\!=\!1$ (noisy).
No associated loss; controls input
augmentation only.

The total training loss is a weighted sum of
$\mathcal{L}_{\text{AE}}$,
$\mathcal{L}_{\text{cone}}$,
$\mathcal{L}_{\text{hm}}$,
$\mathcal{L}_{\text{PILL}}$, and
$\mathcal{L}_{\text{PILLS}}$,
where the physics losses are ramped in
gradually during training.
Full loss definitions and factor details
are given in Appendix~\ref{app:loss_details},
with the total loss in Eq.~\ref{eq:loss_total}.

\subsection{Evaluation}
\label{sec:tracking_measurements}

Tracking performance is measured using
$\ell_1$ errors between predicted and
ground-truth positions at three spatial
scales (56, 112, 224), using three
extraction methods:
\textbf{B}~(bilinear soft-argmax),
\textbf{H}~(hard argmax), and
\textbf{P}~(physics-refined).
The physics module additionally outputs
velocity~$\hat{\mathbf{v}}$ and
bounce detection~$\hat{b}$.
\paragraph{Evaluation protocol.}
Each of the 15 tracking metrics
(B/H/P $\times$ 3~scales, V $\times$ 3,
bounce $\times$ 3) is independently
model-selected on the validation set
(50~sequences).
Test values are reported at the
validation-selected epoch.
The test set is never used for selection.
All results are averaged over $n=4$
replicates (mean $\pm$ std).

\subsection{Factorial Design}
\label{sec:doe_details}

We evaluate all $2^6 = 64$ configurations using a replicated
$2^6$ factorial design with $n=4$ replicates
and $\{-1,+1\}$ contrast coding.
The factorial effect estimate for any main effect
or interaction $K$ is
\[
\hat{\beta}_K
= \frac{1}{n \cdot 2^{k-1}}
  \sum_{j=1}^{n} \sum_{i=1}^{2^k} x_{iK}\, y_{ij},
\quad x_{iK} \in \{-1,+1\},
\]
with $n=4$ replicates and $k=6$ factors.
This equals the difference between the mean response
at the high and low levels of $K$;
negative values indicate improved tracking (lower error).
Each main effect averages over $n \cdot 2^{k-1} = 128$
training runs.

We investigate factorial effects for each of the
nine tracking variables
$y_{ij,m}^{s}$ with extraction method
$m \in \{\mathrm{B}, \mathrm{H}, \mathrm{P}\}$
and scale $s \in \{56, 112, 224\}$,
as well as for two aggregate responses.
The encoder response:
\[
y_{ij,\mathrm{enc,avg}} = \tfrac{1}{3}\bigl(
\mathrm{B56}_{ij} + \mathrm{H56}_{ij}
+ \mathrm{P56}_{ij}\bigr).
\]
The decoder response:
\[
\begin{aligned}
y_{ij,\mathrm{dec,avg}} = \tfrac{1}{6}\bigl(&
\mathrm{B112}_{ij} + \mathrm{B224}_{ij}
+ \mathrm{H112}_{ij} \\
&+ \mathrm{H224}_{ij}
+ \mathrm{P112}_{ij}
+ \mathrm{P224}_{ij}\bigr).
\end{aligned}
\]
Each term is the test $\ell_1$
tracking error at the independently
validation-selected epoch
(Section~\ref{sec:tracking_measurements}).

The total loss is a weighted sum of
$\mathcal{L}_{\text{AE}}$,
$\mathcal{L}_{\text{cone}}$,
$\mathcal{L}_{\text{hm}}$,
$\mathcal{L}_{\text{PILL}}$, and
$\mathcal{L}_{\text{PILLS}}$,
with physics losses ramped in gradually
(Eq.~\ref{eq:loss_total}).
By design, factors A, C, and E interact:
the landmark outputs~(A) gain supervised
semantics when combined with heatmap
supervision~(C) or physics supervision~(E),
yielding AELOS.

\section{Results}\label{Results}
Performance is shown in Sections~\ref{subsec:test_losses_factors} and~\ref{subsec:ablation_study}, with extended discussion of the results provided in Section~\ref{sec:full_factorial_results}.

%First, the networks are first trained with split of roughly the same amount of data, the data is very similar. We increase the amount of noise from no noise $\sigma=0.0$ and finally $\sigma=1.0$. 

\subsection{Test losses of the Factors.}\label{subsec:test_losses_factors}
Tables~\ref{tab:l1-best-enc} and~\ref{tab:l1-best-dec} report the lowest test L1 tracking losses across the nine tracking outputs, shown separately for the two noise settings of factor~F.

% Generated by: python src/scripts/d_min_max_loss_table.py
\begin{table*}[t]
\centering\small
\caption{Encoder ($56^2$) tracking errors ($n=4$ replicates, mean $\pm$ std). Best per column is \underline{underlined}, baseline is \textit{italic}.}
\label{tab:l1-best-enc}
\begin{tabular}{clccc}
\toprule
Row & Config & B56 & H56 & P56 \\
\midrule
\multicolumn{5}{l}{\emph{F=0 ($\sigma=0$)}} \\
 4 & \textit{A0B0C1D0E0F0} & \textit{1.14{\tiny$\pm$0.01}} & \textit{1.13{\tiny$\pm$0.03}} & \textit{1.15{\tiny$\pm$0.01}} \\
\midrule
14 & A0B1C1D1E0F0 & 1.13{\tiny$\pm$0.09} & \underline{1.20{\tiny$\pm$0.09}} & 1.12{\tiny$\pm$0.07} \\
12 & A0B0C1D1E0F0 & \underline{1.12{\tiny$\pm$0.05}} & 1.21{\tiny$\pm$0.06} & \underline{1.12{\tiny$\pm$0.05}} \\
\midrule
\multicolumn{5}{l}{\emph{F=1 ($\sigma=1$)}} \\
36 & \textit{A0B0C1D0E0F1} & \textit{29.50{\tiny$\pm$56.57}} & \textit{30.91{\tiny$\pm$59.55}} & \textit{29.50{\tiny$\pm$56.61}} \\
\midrule
44 & A0B0C1D1E0F1 & \underline{1.20{\tiny$\pm$0.03}} & \underline{1.28{\tiny$\pm$0.04}} & \underline{1.16{\tiny$\pm$0.04}} \\
\bottomrule
\end{tabular}
\end{table*}
\begin{table*}[t]
\centering\small
\caption{Decoder ($112^2$, $224^2$) tracking errors ($n=4$ replicates, mean $\pm$ std). Best per column is \underline{underlined}, baseline is \textit{italic}.}
\label{tab:l1-best-dec}
\begin{tabular}{clcccccc}
\toprule
Row & Config & B112 & B224 & H112 & H224 & P112 & P224 \\
\midrule
\multicolumn{8}{l}{\emph{F=0 ($\sigma=0$)}} \\
 7 & \textit{A1B1C1D0E0F0} & \textit{0.55{\tiny$\pm$0.03}} & \textit{0.41{\tiny$\pm$0.10}} & \textit{0.98{\tiny$\pm$0.06}} & \textit{0.77{\tiny$\pm$0.03}} & \textit{0.62{\tiny$\pm$0.03}} & \textit{0.50{\tiny$\pm$0.12}} \\
\midrule
23 & A1B1C1D0E1F0 & 0.39{\tiny$\pm$0.03} & \underline{0.31{\tiny$\pm$0.01}} & 1.04{\tiny$\pm$0.02} & \underline{0.96{\tiny$\pm$0.03}} & 0.45{\tiny$\pm$0.02} & \underline{0.38} \\
31 & A1B1C1D1E1F0 & \underline{0.38{\tiny$\pm$0.03}} & 0.34{\tiny$\pm$0.04} & \underline{0.97{\tiny$\pm$0.07}} & 1.01{\tiny$\pm$0.03} & \underline{0.42} & 0.38{\tiny$\pm$0.02} \\
\midrule
\multicolumn{8}{l}{\emph{F=1 ($\sigma=1$)}} \\
39 & \textit{A1B1C1D0E0F1} & \textit{3.40{\tiny$\pm$1.88}} & \textit{2.16{\tiny$\pm$1.63}} & \textit{0.98{\tiny$\pm$0.04}} & \textit{0.89{\tiny$\pm$0.05}} & \textit{2.88{\tiny$\pm$1.29}} & \textit{1.98{\tiny$\pm$2.19}} \\
\midrule
55 & A1B1C1D0E1F1 & \underline{0.41{\tiny$\pm$0.02}} & \underline{0.35{\tiny$\pm$0.02}} & \underline{1.07{\tiny$\pm$0.07}} & \underline{1.05{\tiny$\pm$0.06}} & 0.46{\tiny$\pm$0.01} & \underline{0.41{\tiny$\pm$0.01}} \\
63 & A1B1C1D1E1F1 & 0.41{\tiny$\pm$0.02} & 0.37{\tiny$\pm$0.02} & 1.08{\tiny$\pm$0.13} & 1.11{\tiny$\pm$0.06} & \underline{0.45{\tiny$\pm$0.02}} & 0.41{\tiny$\pm$0.02} \\
\bottomrule
\end{tabular}
\end{table*}

% Inference heatmap figure — smaller and moved next to Fig. 4 in Section 4.3.

For both noise conditions, configurations that include the supervised physics-informed loss (PILLS, factor~E) appear most frequently among the rows achieving the lowest losses across tracking variables.

In the noise-free setting ($F{=}0$), rows~23 and~31 (both A1B1C1E1) achieve the lowest decoder errors for bilinear and physics-refined outputs, with sub-pixel accuracy (${\leq}0.42$\,px at scale~112).
Under noisy conditions ($F{=}1$), row~55 (A1B1C1D0E1F1) attains the best bilinear and physics-refined decoder errors, demonstrating that PILLS maintains sub-pixel accuracy even under $\sigma{=}1$ noise.
Row~39 (the baseline, A1B1C1D0E0F1) retains the lowest hard-argmax errors at scales~112 and~224, consistent with hard-argmax being less sensitive to heatmap shape but limited to integer resolution.

At the lowest resolution (B56 and P56), performance degrades compared to higher resolutions, indicating that bilinear upsampling combined with physics-module prediction provides an insufficient learning signal at the current parameter settings.%, despite stable behavior at finer resolutions.

\subsection{Factorial effects.}\label{subsec:ablation_study}
The factorial effects are shown per tracking variable in Tables~\ref{tab:effects-enc} (encoder) and~\ref{tab:effects-dec} (decoder), with the 10 largest effects ranked by average magnitude.
% Generated by: python src/scripts/d_contrast_encoder_decoder.py
\begin{table}[t]
\centering\small
\caption{Encoder factorial effects ($n=4$ replicates). Negative = reduces error.}
\label{tab:effects-enc}
\begin{tabular}{lrrrr}
\toprule
Factor & B56 & H56 & P56 & Avg \\
\midrule
C & -24.89 & -25.10 & -23.88 & \underline{-24.62} \\
AB & -15.36 & -16.12 & -14.84 & -15.44 \\
BC & -14.21 & -15.03 & -12.94 & -14.06 \\
AC & +11.10 & +11.73 & +11.10 & +11.31 \\
A & +10.39 & +10.88 & +10.84 & +10.71 \\
ABC & -10.35 & -10.82 & -10.26 & -10.48 \\
DE & +6.13 & +6.32 & +6.97 & +6.47 \\
B & -6.05 & -6.25 & -4.97 & -5.76 \\
ABD & -5.52 & -5.80 & -5.31 & -5.55 \\
CE & +4.62 & +4.62 & +4.82 & +4.69 \\
\bottomrule
\end{tabular}
\end{table}
\begin{table*}[t]
\centering\small
\caption{Decoder factorial effects ($n=4$ replicates). Negative = reduces error.}
\label{tab:effects-dec}
\begin{tabular}{lrrrrrrr}
\toprule
Factor & B112 & B224 & H112 & H224 & P112 & P224 & Avg \\
\midrule
A & -21.26 & -26.98 & -16.91 & -20.07 & -22.11 & -27.50 & \underline{-22.47} \\
C & -14.50 & -10.50 & -32.60 & -23.15 & -13.48 & -9.42 & -17.27 \\
AC & -11.79 & -10.51 & -12.63 & -4.46 & -10.31 & -9.34 & -9.84 \\
E & -9.15 & -11.23 & -3.79 & -5.35 & -10.75 & -13.14 & -8.90 \\
AE & -8.64 & -11.15 & -2.89 & -4.11 & -9.38 & -12.24 & -8.07 \\
AB & -0.36 & +0.36 & -22.40 & -18.26 & +1.47 & +1.50 & -6.28 \\
BC & -5.12 & -5.88 & -6.73 & -3.77 & -5.88 & -5.63 & -5.50 \\
ABC & -2.41 & -5.89 & -8.66 & -6.80 & -2.69 & -5.56 & -5.34 \\
CE & +2.09 & +5.15 & +8.92 & +6.79 & +2.50 & +6.49 & +5.32 \\
ACE & +1.02 & +4.67 & +2.78 & +7.54 & +1.78 & +5.23 & +3.84 \\
\bottomrule
\end{tabular}
\end{table*}

The dominant main effects are associated with the physics-informed tracking (PILLS) factor (E) of the model for both the encoder and decoder. In particular, the supervised physics loss (E) and decoder (A) exhibit the largest negative effects on the decoder, indicating a substantial reduction in the test error when these main effects are enabled for the decoder and also combined in a second order interaction. 

Input noise augmentation (F) shows a strong positive effect as expected, suggesting that noise increases the difficulty of the tracking task. Especially, the main effect E has a negative effect, but also on all the higher order interactions it is part of. 
Surprisingly, Factor C main effect has positive values for the encoder tracking similar to Factor A main effects, but negative as expected for the decoder. Opposite C, Factor D has positive values in the decoder but negative values in the encoder, and it is part of four beneficial higher order interaction terms. 
\subsection{Velocity and bounce prediction.}
\label{subsec:velocity_bounce}
\begin{figure}[t]
\centering
\begin{subfigure}[c]{0.44\columnwidth}
  \centering
  \includegraphics[width=\linewidth]{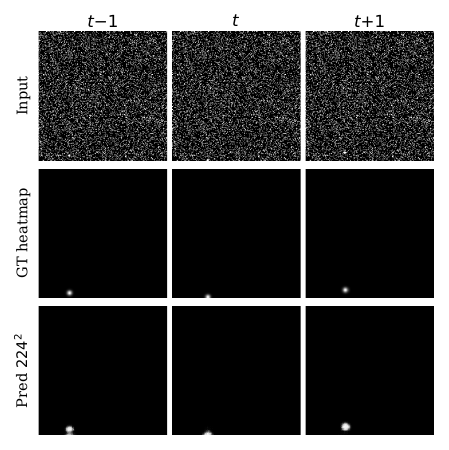}
  \caption{Inference under noise.}
  \label{fig:inference}
\end{subfigure}
\hfill
\begin{subfigure}[c]{0.52\columnwidth}
  \centering
  \begin{subfigure}[b]{0.75\linewidth}
    \centering
    \includegraphics[width=\linewidth]{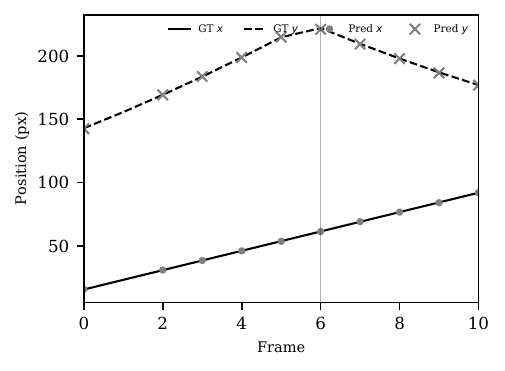}
    \caption{Position (frames 0--10).}
    \label{fig:position}
  \end{subfigure}\\[0.2em]
  \begin{subfigure}[b]{0.75\linewidth}
    \centering
    \includegraphics[width=\linewidth]{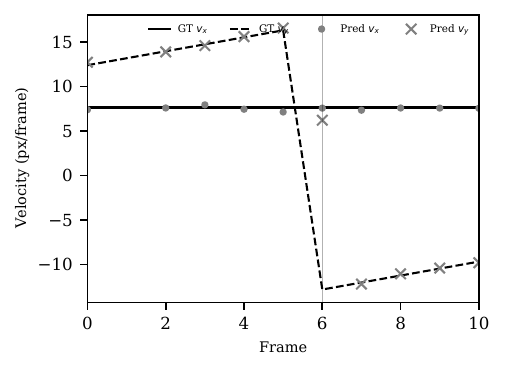}
    \caption{Velocity ($v_y$ flip at frame~6 = bounce).}
    \label{fig:velocity}
  \end{subfigure}
\end{subfigure}
\caption{Qualitative results on test video~78, Row~55 (A1B1C1E1F1, $\sigma=1$, seed~100).
\textbf{(a)} Input frame, ground-truth heatmap, and predicted decoder heatmap at $224^2$ for frames 5--7 under noisy conditions --- the ball is nearly invisible in the input, yet the predicted heatmap concentrates on the correct location.
\textbf{(b)}~Predicted ball position (bilinear and physics-refined) matches the ground-truth parabolic trajectory over the full 11-frame window.
\textbf{(c)}~Predicted velocity captures the abrupt $v_y$ sign flip at frame~6, which corresponds to a wall bounce in the ground-truth trajectory.
All physical outputs are produced by the differentiable physics module in a single forward pass.}
\label{fig:pos_vel}
\end{figure}

A key advantage of the physics-informed approach is that the model outputs velocity estimates and bounce detection from a single forward pass. Table~\ref{tab:velocity} reports velocity prediction errors across configurations, and Figure~\ref{fig:pos_vel} shows qualitative inference results on test video~78 (Row~55, A1B1C1E1F1). Only configurations with both the multi-scale decoder (A=1) and noise bottleneck (B=1) achieve low errors at all three scales; without these, the 112 and 224 scale heatmaps collapse to ${\sim}57$~px error while the 56-scale encoder heatmap remains functional.

% Generated by: python src/scripts/d_generate_tables_aggregated.py
\begin{table*}[t]
\centering\small
\caption{Velocity prediction errors on the test set ($n=4$ replicates, mean $\pm$ std). Best per column and $F$ is \underline{underlined}, baseline is \textit{italic}.}
\label{tab:velocity}
\begin{tabular}{clccc}
\toprule
Row & Config & V56 & V112 & V224 \\
\midrule
\multicolumn{5}{l}{\emph{F=0 ($\sigma=0$)}} \\
 7 & \textit{A1B1C1D0E0F0} & \textit{1.19{\tiny$\pm$0.04}} & \textit{0.93{\tiny$\pm$0.03}} & \textit{0.70{\tiny$\pm$0.09}} \\
\midrule
12 & A0B0C1D1E0F0 & \underline{1.14{\tiny$\pm$0.05}} & 8.17 & 8.17 \\
31 & A1B1C1D1E1F0 & 1.27{\tiny$\pm$0.05} & \underline{0.62{\tiny$\pm$0.03}} & \underline{0.60{\tiny$\pm$0.04}} \\
\midrule
\multicolumn{5}{l}{\emph{F=1 ($\sigma=1$)}} \\
39 & \textit{A1B1C1D0E0F1} & \textit{62.28{\tiny$\pm$46.57}} & \textit{2.17{\tiny$\pm$0.61}} & \textit{1.24{\tiny$\pm$0.42}} \\
\midrule
47 & A1B1C1D1E0F1 & \underline{1.27{\tiny$\pm$0.05}} & 6.89{\tiny$\pm$0.12} & 7.15{\tiny$\pm$0.15} \\
63 & A1B1C1D1E1F1 & 13.13{\tiny$\pm$14.62} & \underline{0.69{\tiny$\pm$0.06}} & \underline{0.66{\tiny$\pm$0.06}} \\
\bottomrule
\end{tabular}
\end{table*}

\section{Conclusion and discussion}
Using a controlled tracking task and a replicated factorial experimental design ($n=4$ replicates), we studied the effect of physics-informed landmark constraints on tracking performance in multiple architectural choices, supervision levels, and noise conditions.
Overall, the results indicate that physics-informed landmark losses provide a principled mechanism for injecting physical structure into learning-based tracking models, with supervised physics-informed training (PILLS) consistently yielding the lowest or near-lowest test tracking errors across noise conditions. While hard argmax (H) is the standard detection, bilinearly interpolated predictions at B112 and B224—and their corresponding physics-refined outputs (P112 and P224)—outperform hard argmax in several cases, likely due to improved sub-pixel localization. At B56 and P56, the coarse spatial resolution leads to degraded performance compared to H56.

Although reconstruction quality remains limited in this simple setup, the primary objective of this study is accurate landmark tracking. Beyond tracking, the proposed physics-informed landmark framework can naturally extend to generative settings since the approach enables trajectory generation that is both visually plausible and physically consistent.
A key advantage of the physics-informed approach is that the model outputs not only refined positions but also velocity estimates and bounce detection from a single forward pass --- physical state predictions that are not available from standard heatmap-based tracking methods.

Limitations. The current experiments track a single particle; multi-particle tracking with crossing trajectories is not addressed and is left for future work. All experiments use synthetic data, and validation on real-world video remains future work. The method depends on correct physical priors --- if the assumed dynamics (gravity, bounce model) do not match the true system, performance may degrade.

\section*{Acknowledgements}
This work was supported by a research grant (VIL77978) from Villum Fonden, as part of the Villum Experiment Programme project ``Scientific Machine Learning for Advancing the Understanding of the 4D Microstructural Evolution in Metals (SciML4D).''

\bibliographystyle{plainnat}
% Single source of truth: Zotero-exported clean library
% (symlinked to "My Library_clean.bib" to avoid the space
% in the filename, which bibtex cannot parse).
\bibliography{library_clean}

%%%%%%%%%%%%%%%%%%%%%%%%%%%%%%%%%%%%%%%%%%%%%%%%%%%%%%%%%%%%%%%%%%%%%%%%%%%%%%%
%%%%%%%%%%%%%%%%%%%%%%%%%%%%%%%%%%%%%%%%%%%%%%%%%%%%%%%%%%%%%%%%%%%%%%%%%%%%%%%
% APPENDIX
%%%%%%%%%%%%%%%%%%%%%%%%%%%%%%%%%%%%%%%%%%%%%%%%%%%%%%%%%%%%%%%%%%%%%%%%%%%%%%%
%%%%%%%%%%%%%%%%%%%%%%%%%%%%%%%%%%%%%%%%%%%%%%%%%%%%%%%%%%%%%%%%%%%%%%%%%%%%%%%
\newpage
\appendix
\onecolumn

\section{PIT: Encoder-Decoder Architecture, Losses and Factors}
\label{PIT_network}
This section describes the full PIT architecture following Table~\ref{tab:pit-architecture}, Figure~\ref{fig:network_architecture}, and Algorithm~\ref{alg:pit}, with each factor and its associated loss introduced where it enters the computational graph.

\begin{algorithm*}[t]
\caption{PIT: Training procedure. Factors $A, B, C, D, E, F \in \{0,1\}$ control which components and losses are active.}
\label{alg:pit}
\begin{algorithmic}[1]
\REQUIRE Training set; known physics $g, e, \Delta t$; factors $A, B, C, D, E, F \in \{0,1\}$
\ENSURE $\hat{\mathbf{p}}_\tau^{s}$, $\hat{\mathbf{v}}_\tau$, $\hat{b}_\tau$ at $s \in \{56, 112, 224\}$
\FOR{epoch $= 1, \ldots, N$}
\FOR{each 3-frame window $(I_{t-1}, I_t, I_{t+1})$ in training set}
\STATE \textbf{Encode} each frame with shared weights:
\FOR{$\tau \in \{t\!-\!1, t, t\!+\!1\}$}
  \STATE $\mathbf{z}_\tau = \text{Enc}(I_\tau)$ \hfill $\triangleright$ $56\!\times\!56$ features
  \STATE $H_\tau = \text{CenterHead}(\mathbf{z}_\tau)$ \hfill $\triangleright$ A: tracking heatmap
  \STATE $R_\tau = \text{ResidualHead}(\mathbf{z}_\tau)$ \hfill $\triangleright$ B: residual
\ENDFOR
\STATE \textbf{Fuse:} $\mathbf{z}_{\text{fused}} = \text{Conv}_{1\times1}([H_1,\ldots,H_T,\,R_1,\ldots,R_T])$
\STATE $H^{56} = \sigma(\mathbf{z}_{\text{fused}}[1{:}T])$; \; $R^{56} = \mathbf{z}_{\text{fused}}[T{+}1{:}]$
\STATE \textbf{Track at $56^2$:} $\hat{\mathbf{p}}_\tau^{56} = \text{coarse-to-fine}(H^{56}_\tau)$
\STATE \textbf{Physics at $56^2$} (if $D{=}1$ or $E{=}1$): $\tilde{\mathbf{p}}, \hat{\mathbf{v}}, \hat{b} = f(\hat{\mathbf{p}}^{56};\, g, e, \Delta t)$ \hfill $\triangleright$ \texttt{torch.where}
\STATE \textbf{Decode $56 \!\to\! 112$}: detach $H^{56}, R^{56}$ if $A{=}0$
\STATE \quad $H^{112} = \sigma(\text{DecCenter}_{112}([\text{Up}(D^{56}),\, \uparrow\!H^{56}]))$
\STATE \quad Track: $\hat{\mathbf{p}}_\tau^{112} = \text{biquadratic}(H^{112}_\tau)$; physics at $112^2$
\STATE \textbf{Decode $112 \!\to\! 224$}:
\STATE \quad $H^{224} = \sigma(\text{DecCenter}_{224}([\text{Up}(D^{112}),\, \uparrow\!H^{112},\, \uparrow\!\uparrow\!H^{56}]))$
\STATE \quad Track: $\hat{\mathbf{p}}_\tau^{224} = \text{bicubic}(H^{224}_\tau)$; physics at $224^2$
\STATE \textbf{Reconstruct:} $\hat{I} = \text{DecImg}(D^{224})$
\STATE \textbf{Losses} at all scales $s \in \{56, 112, 224\}$:
\STATE \quad \textbf{if} $A{=}1$: $\mathcal{L}_{\text{recon}} = \text{BCE}(\hat{I}, I) + \mathcal{L}_{\text{cone}}$
\STATE \quad \textbf{if} $C{=}1$: $\mathcal{L}_{\text{hm}}^s$ = focal heatmap loss
\STATE \quad \textbf{if} $D{=}1$: $\mathcal{L}_{\text{PILL}}^s = \| \tilde{\mathbf{p}}^s_{t+1} - \hat{\mathbf{p}}^s_{t+1} \|_1$
\STATE \quad \textbf{if} $E{=}1$: $\mathcal{L}_{\text{PILLS}}^s = \| \hat{\mathbf{p}}^s \!-\! \mathbf{p}^* \|_1 + \| \hat{\mathbf{v}}^s \!-\! \mathbf{v}^* \|_1 + \text{BCE}(\hat{b}^s, b^*)$
\STATE \textbf{Update:} $\theta \!\leftarrow\! \theta - \eta\,\nabla_\theta\!\big(\mathcal{L}_{\text{recon}} + \sum_s \mathcal{L}_{\text{hm}}^s + w_D \mathcal{L}_{\text{PILL}}^s + w_E \mathcal{L}_{\text{PILLS}}^s\big)$
\ENDFOR
\ENDFOR
\end{algorithmic}
\end{algorithm*}

\begin{table*}[t!]
\centering
\footnotesize
\caption{Complete PIT architecture: layer specifications, tensor shapes, and factor dependencies.
ResBlock($c$) = Conv$_{3\times3}$--BN--ReLU--Conv$_{3\times3}$--BN + skip $\to$ ReLU.
$C_r\!=\!6$ residual channels, $T\!=\!3$ frames, softargmax $\beta\!=\!15$.
Sigmoid activation is applied to center heatmaps at all spatial scales.
Factor~F (noise) is applied to input; cone center is GT (C/E=1) or predicted (C=E=0).}
\label{tab:pit-architecture}
\begin{tabular}{lllll}
\toprule
\textbf{Stage / Output} & \textbf{Layer} & \textbf{In$\to$Out} & \textbf{Shape} & \textbf{Factor} \\
\midrule

\multicolumn{5}{l}{\emph{Input}} \\
Frames $I_{t-1}, I_t, I_{t+1}$ & Noise augmentation & -- & $(B,3,224,224)$ & F \\

\midrule
\multicolumn{5}{l}{\emph{Encoder (shared weights, $224 \to 56$)}} \\
& Conv$_{3\times3}$--BN--ReLU & $1 \to 32$ & $(B,3,32,224,224)$ & -- \\
& ResBlock(32) & $32 \to 32$ & $(B,3,32,224,224)$ & -- \\
& MaxPool $2\times2$ & -- & $(B,3,32,112,112)$ & -- \\
& Conv$_{3\times3}$--BN--ReLU & $32 \to 64$ & $(B,3,64,112,112)$ & -- \\
& ResBlock(64) & $64 \to 64$ & $(B,3,64,112,112)$ & -- \\
& MaxPool $2\times2$ & -- & $(B,3,64,56,56)$ & -- \\

\midrule
\multicolumn{5}{l}{\emph{Split bottleneck}} \\
$H^{56}$ (CenterHead) & Conv$_{1\times1}$ & $64 \to 1$ & $(B,3,1,56,56)$ & -- \\
$R^{56}$ (ResidualHead) & Conv$_{3\times3}$--ReLU & $64 \to 6$ & $(B,3,6,56,56)$ & A / B \\
LatentFuse & Conv$_{1\times1}$--ReLU & $21 \to 21$ & $(B,21,56,56)$ & -- \\

\midrule
\multicolumn{5}{l}{\emph{Tracking readout ($56^2$)}} \\
$H^{56}_{\mathrm{dec}}=\sigma(H^{56})$ & Sigmoid & -- & $(B,3,56,56)$ & C$_{56}$ \\
$\hat{\mathbf{c}}^{56}_{\mathrm{bil}}$ & Coarse-to-fine expectation & -- & $(B,3,2)$ & -- \\
$\hat{\mathbf{c}}^{56}_{\mathrm{arg}}$ & Hard argmax (eval only) & -- & $(B,3,2)$ & -- \\

\midrule
\multicolumn{5}{l}{\emph{Physics module ($56^2$, applied at all scales)}} \\
$\hat{\mathbf{p}}^{56}_{\mathrm{phys}}$ & Velocity-Verlet + \texttt{torch.where} & -- & $(B,3,2)$ & D$_{56}$ / E$_{56}$ \\
$\hat{\mathbf{v}}^{56}, \hat{b}^{56}$ & Velocity, bounce & -- & $(B,3,2)$, $(B,3)$ & E$_{56}$ \\

\midrule\midrule
\multicolumn{5}{l}{\emph{Decoder $56 \to 112$}} \\
dec\_conv0 & Conv$_{3\times3}$ & $21 \to 24$ & $(B,24,56,56)$ & -- \\
dec\_up1 & ConvTranspose$_{2\times2}$ s2 & $24 \to 16$ & $(B,16,112,112)$ & -- \\
& + skip: concat $\uparrow\!H^{56}$ & $16\!+\!3 = 19$ & $(B,19,112,112)$ & A \\
$H^{112}$ (dec\_center\_2) & Conv$_{1\times1}$ + $\sigma$ & $19 \to 3$ & $(B,3,112,112)$ & C$_{112}$ \\
$\hat{\mathbf{c}}^{112}_{\mathrm{bil}}$ & Biquadratic expectation & -- & $(B,3,2)$ & -- \\
$\hat{\mathbf{c}}^{112}_{\mathrm{arg}}$ & Hard argmax (eval only) & -- & $(B,3,2)$ & -- \\
$\hat{\mathbf{p}}^{112}, \hat{\mathbf{v}}^{112}, \hat{b}^{112}$ & Physics module & -- & $(B,3,2)$, $(B,3)$ & D$_{112}$ / E$_{112}$ \\

\midrule
\multicolumn{5}{l}{\emph{Decoder $112 \to 224$}} \\
dec\_up2 & ConvTranspose$_{2\times2}$ s2 & $19 \to 8$ & $(B,8,224,224)$ & -- \\
& + skip: $\uparrow\!H^{112}$, $\uparrow\!H^{56}$ & $8\!+\!3\!+\!3 = 14$ & $(B,14,224,224)$ & A \\
$H^{224}$ (dec\_center\_4) & Conv$_{1\times1}$ + $\sigma$ & $14 \to 3$ & $(B,3,224,224)$ & C$_{224}$ \\
$\hat{\mathbf{c}}^{224}_{\mathrm{bil}}$ & Bicubic expectation & -- & $(B,3,2)$ & -- \\
$\hat{\mathbf{c}}^{224}_{\mathrm{arg}}$ & Hard argmax (eval only) & -- & $(B,3,2)$ & -- \\
$\hat{\mathbf{p}}^{224}, \hat{\mathbf{v}}^{224}, \hat{b}^{224}$ & Physics module & -- & $(B,3,2)$, $(B,3)$ & D$_{224}$ / E$_{224}$ \\

\midrule
\multicolumn{5}{l}{\emph{Reconstruction}} \\
$\hat{I}$ (dec\_img) & Conv$_{1\times1}$ & $14 \to 3$ & $(B,3,224,224)$ & A / B \\
$\mathcal{L}_{\text{cone}}$ & Cone-guided recon loss & -- & scalar & A \\

\bottomrule
\end{tabular}
\end{table*}

%%%%%%%%%%%%%%%%%%%%%%%%%%%%%%%%%%%%%%%%%%%%%%%%%%%%%%%%%%%%

\subsection{Input and Factor~F (Noise condition)}

The network receives three consecutive grayscale frames $I_{t-1}, I_t, I_{t+1}$ ($1{\times}224{\times}224$), with Factor~F controlling input noise augmentation ($\sigma{=}0$ clean, $\sigma{=}1$ noisy).

\subsection{Encoder and split bottleneck}

A shared convolutional encoder processes each frame independently, downsampling from $224{\times}224$ to $56{\times}56$ through two stages of Conv--ResBlock--Pool (Table~\ref{tab:pit-architecture}).
The encoder features then split into two parallel heads:
$H^{56}$ (\textbf{CenterHead}, Conv$_{1\times1}$), a single-channel heatmap for tracking; and
$R^{56}$ (\textbf{ResidualHead}, Conv$_{3\times3}$--ReLU), $C_r{=}6$ residual channels capturing appearance and noise.
These are concatenated across the $T{=}3$ frames ($3{\times}(1{+}6){=}21$ channels) and fused via a $1{\times}1$ convolution (LatentFuse).
The first $T{=}3$ channels of the fused representation are passed through a sigmoid to produce center heatmaps $H^{56}_{\mathrm{dec}} = \sigma(z_{1:T})$; the remaining $3C_r{=}18$ channels form the residual $R^{56}_{\mathrm{dec}}$.

\subsection{Factor~A: Autoencoder with landmark outputs (AELO/AELOS)}\label{loss:ae}\label{loss:cone}

Factor~A controls whether landmark heatmaps are passed into the decoder via skip connections, coupling tracking with reconstruction.

In the \textbf{AELO} setting (without Factors~C and~E), no explicit tracking supervision is applied to the decoder-level outputs.
The decoder is trained purely through reconstruction, where local cone masks are generated from the \emph{predicted} heatmap peaks and used to guide the reconstruction loss around the ball location.
This allows the encoder to be updated indirectly via reconstruction gradients, without direct supervision of higher-resolution tracking.
In the \textbf{AELOS} setting, heatmap supervision~(C) and/or physics supervision~(E) are additionally enabled, resulting in jointly supervised tracking and reconstruction across all decoder scales.

When $A{=}1$, gradients flow from the decoder back to the encoder through the skip-connected heatmaps.
When $A{=}0$, the heatmaps are detached from the computational graph before entering the decoder.

The decoder reconstructs the input frame: $\hat{I} = \mathrm{DecImg}(D^{224}) \in \mathbb{R}^{3 \times 224 \times 224}$.
The reconstruction loss (always active) is:
\begin{equation}
\begin{aligned}
\mathcal{L}_{\mathrm{AE}}
&=
-\frac{1}{N}
\sum_{n=1}^{N}
\Big[
I^{(n)} \log \sigma(\hat{I}^{(n)}) \\
&\quad+
\big(1 - I^{(n)}\big)
\log\!\big(1 - \sigma(\hat{I}^{(n)})\big)
\Big],
\end{aligned}
\label{eq:loss_ae}
\end{equation}
where $\hat{I}$ is the decoder output in logit space and $\sigma(\cdot)$ is applied implicitly (\texttt{BCEWithLogitsLoss}).

To focus the reconstruction near the tracked object, a cone-guided loss weights the error using a Gaussian mask
$C_{b}(x,y) = \exp\!\left(
-\frac{(x - c_x)^2 + (y - c_y)^2}{2\sigma^2}
\right)$
with $\sigma = 3r$, $r\!=\!2$~px,
normalized to unit peak:
\begin{equation}
\mathcal{L}_{\mathrm{cone}}
=
\frac{1}{B}
\sum_{b=1}^{B}
\frac{1}{HW}
\sum_{x,y}
\left|
\hat{I}_{b}(x,y) - I_{b}(x,y)
\right|
\tilde{C}_{b}(x,y).
\label{eq:loss_cone}
\end{equation}
The cone center corresponds to the predicted ball location (AELO) or ground-truth location (AELOS when C or E is active).

\subsection{Factor~B: Noise bottleneck}

Factor~B controls whether the non-tracking residual channels $R^{56}_{\mathrm{dec}}$ are passed into the decoder for background and noise reconstruction.
When $B{=}1$, these channels provide an unstructured pathway and gradients from $\mathcal{L}_{\mathrm{AE}}$ flow back to the encoder through them.
When $B{=}0$, the residual channels are detached, isolating the encoder from reconstruction gradients through this pathway.

\subsection{Factor~C: Heatmap supervision}\label{loss:hm}

Factor~C enables explicit landmark supervision via focal heatmap loss, similar to CenterNet~\cite{duanCenterNetKeypointTriplets2019b}.
At each spatial scale ($56$, $112$, $224$), the network predicts a center heatmap whose peak corresponds to the object location.
Ground-truth landmarks are encoded as Gaussian target heatmaps $\tilde{H}$:
\begin{equation}
\begin{aligned}
\mathcal{L}_{\mathrm{hm}} =
- \frac{1}{|\mathcal{P}|}
\sum_{i,j}
\Big[
&\mathbb{1}_{\tilde{H}_{ij}>0.5}
(1\!-\!H_{ij})^{2}\log H_{ij} \\
&+
(1\!-\!\tilde{H}_{ij})^{4}
H_{ij}^{2}\log(1\!-\!H_{ij})
\Big],
\end{aligned}
\label{eq:loss_hm}
\end{equation}
where $|\mathcal{P}|$ is the number of positive pixels.
This loss promotes sharp, localized heatmap peaks while suppressing background responses.
When $C{=}0$, no direct heatmap supervision is applied; landmark locations are instead inferred implicitly through reconstruction~(A) and physics losses~(D, E).

\subsection{Tracking readout and decoder}

Landmark coordinates are extracted from $H^{56}_{\mathrm{dec}}$ using two methods:
(1)~a differentiable \emph{coarse-to-fine expectation} (hard argmax followed by local soft-argmax in an $r{=}3$ pixel neighborhood; Section~\ref{app:bilinear_expectation}), yielding sub-pixel positions $\hat{\mathbf{c}}^{56}_{\mathrm{bil}}$; and
(2)~a \emph{hard argmax} $\hat{\mathbf{c}}^{56}_{\mathrm{arg}}$ used only at evaluation.
At decoder scales, different expectation operators provide increasingly smooth sub-pixel refinement: biquadratic at $112$ and bicubic at $224$.

The decoder input is $D^{56} = [H^{56}_{\mathrm{dec}},\, R^{56}_{\mathrm{dec}}]$ ($21$ channels).
At each decoder stage, upsampled encoder heatmaps are injected as skip connections (controlled by Factor~A), a refined heatmap is predicted (supervised by Factor~C), and sub-pixel positions are extracted and passed to the physics module (Factors~D/E):

\paragraph{$56 \to 112$:}
Conv$_{3\times3}$ ($21{\to}24$), ConvTranspose ($24{\to}16$), skip-concat with $\uparrow\!H^{56}$ ($16{+}3{=}19$).
Refined heatmap: $H^{112} = \sigma(\mathrm{DecCenter}_{112}(D^{112}))$.
Positions extracted via biquadratic expectation.

\paragraph{$112 \to 224$:}
ConvTranspose ($19{\to}8$), skip-concat with $\uparrow\!H^{112}$ and $\uparrow\!\uparrow\!H^{56}$ ($8{+}3{+}3{=}14$).
Final heatmap: $H^{224} = \sigma(\mathrm{DecCenter}_{224}(D^{224}))$.
Positions extracted via bicubic expectation.
Reconstruction: $\hat{I} = \mathrm{DecImg}(D^{224})$ ($14{\to}3$ channels).

\subsection{Returned outputs}
The forward pass returns center heatmaps, tracking positions (soft and hard), physics predictions, and the reconstructed image at all three scales:
\begin{equation}
\Big(
H^{s},\,
\hat{\mathbf{c}}^{s}_{\mathrm{bil}},\,
\hat{\mathbf{c}}^{s}_{\mathrm{arg}},\,
\hat{\mathbf{p}}^{s}_{\mathrm{phys}},\,
\hat{\mathbf{v}}^{s},\,
\hat{b}^{s}
\Big)_{s\in\{56,112,224\}},\;
\hat{I}.
\end{equation}

%%%%%%%%%%%%%%%%%%%%%%%%%%%%%%%%%%%%%%%%%%%%%%%%%%%%%%%%%%%%

\subsection{Soft-Argmax Expectation Operators}
\label{app:bilinear_expectation}

All soft-argmax operators share the same preprocessing: given a predicted heatmap $H \in \mathbb{R}^{H \times W}$,
negative values are suppressed via $\tilde{H}_{ij} = \max(H_{ij}, 0)$.
Each operator then computes sub-pixel coordinates $(x,y)$ as a weighted centroid of $\tilde{H}$.
Different operators are used at different spatial scales to balance localization precision with robustness.

\subsubsection{Bilinear expectation (baseline)}
The standard bilinear expectation computes the global weighted centroid:
\begin{equation}
M = \textstyle\sum_{i,j} \tilde{H}_{ij} + \varepsilon,
\quad
x = \frac{1}{M}\textstyle\sum_{i,j} j\,\tilde{H}_{ij},
\quad
y = \frac{1}{M}\textstyle\sum_{i,j} i\,\tilde{H}_{ij}.
\end{equation}
This is fully differentiable but sensitive to distant activations.

\subsubsection{Coarse-to-fine expectation ($56{\times}56$)}
At the coarse encoder scale, a two-step approach is used.
First, the hard argmax finds the peak $(x_c, y_c)$ (detached from gradients).
Second, a bilinear expectation is computed within a local window of radius $r_w{=}3$ pixels:
\begin{equation}
\mathcal{N} = \{(i,j) : |j - x_c| \leq r_w \;\text{and}\; |i - y_c| \leq r_w\},
\end{equation}
\begin{equation}
x = \frac{\sum_{(i,j) \in \mathcal{N}} \tilde{H}_{ij}\, j}
         {\sum_{(i,j) \in \mathcal{N}} \tilde{H}_{ij} + \varepsilon},
\qquad
y = \frac{\sum_{(i,j) \in \mathcal{N}} \tilde{H}_{ij}\, i}
         {\sum_{(i,j) \in \mathcal{N}} \tilde{H}_{ij} + \varepsilon}.
\end{equation}
Gradients flow through the second step only, making this robust to multi-modal heatmaps at coarse resolution.

\subsubsection{Biquadratic expectation ($112{\times}112$)}
A two-pass refinement: first compute the bilinear centroid $(\bar{x}, \bar{y})$,
then reweight the heatmap with a quadratic kernel that suppresses distant pixels:
\begin{equation}
w_{ij} = \max\!\Big(1 - \frac{(j - \bar{x})^2 + (i - \bar{y})^2}{4},\; 0\Big),
\end{equation}
\begin{equation}
x = \frac{\sum_{i,j} w_{ij}\,\tilde{H}_{ij}\, j}
         {\sum_{i,j} w_{ij}\,\tilde{H}_{ij} + \varepsilon},
\qquad
y = \frac{\sum_{i,j} w_{ij}\,\tilde{H}_{ij}\, i}
         {\sum_{i,j} w_{ij}\,\tilde{H}_{ij} + \varepsilon}.
\end{equation}
The quadratic falloff provides a smooth transition between local and global weighting.

\subsubsection{Bicubic expectation ($224{\times}224$)}
Similar two-pass approach, but with separable cubic kernels for a wider support:
\begin{equation}
w_x = \max\!\Big(1 - \frac{|j - \bar{x}|^3}{8},\; 0\Big),
\quad
w_y = \max\!\Big(1 - \frac{|i - \bar{y}|^3}{8},\; 0\Big),
\quad
w_{ij} = w_x \cdot w_y,
\end{equation}
\begin{equation}
x = \frac{\sum_{i,j} w_{ij}\,\tilde{H}_{ij}\, j}
         {\sum_{i,j} w_{ij}\,\tilde{H}_{ij} + \varepsilon},
\qquad
y = \frac{\sum_{i,j} w_{ij}\,\tilde{H}_{ij}\, i}
         {\sum_{i,j} w_{ij}\,\tilde{H}_{ij} + \varepsilon}.
\end{equation}
The cubic falloff is smoother and wider than the quadratic kernel, suitable for the full-resolution heatmaps where the peak is well-resolved.
All three operators are fully differentiable (except the hard-argmax step in coarse-to-fine, which is detached).

%%%%%%%%%%%%%%%%%%%%%%%%%%%%%%%%%%%%%%%%%%%%%%%%%%%%%%%%%%%%

\subsection{Hard Arg-max Landmark Extraction}
\label{app:hard_argmax}

In addition to differentiable landmark extraction, we also compute discrete
landmark locations using a hard argmax operator.
This operation extracts the spatial location of the maximum activation in a
predicted heatmap and is used exclusively for evaluation and diagnostic purposes.

Let $H \in \mathbb{R}^{H \times W}$ denote a predicted center heatmap.
The hard argmax landmark $\mathbf{c} = (x,y)$ is defined as
\begin{equation}
(x,y)
=
\arg\max_{(i,j)} H_{ij}.
\end{equation}
In practice, the heatmap is flattened into a vector, and the index of the maximum
value is identified.
This index is then converted back to two-dimensional coordinates using integer
division and modulo operations,
\begin{equation}
y = \left\lfloor \frac{k}{W} \right\rfloor,
\qquad
x = k \bmod W,
\end{equation}
where $k$ denotes the index of the maximum value in the flattened heatmap and $W$
is the heatmap width.

The resulting coordinates are returned in heatmap coordinate space as
$\mathbf{c} \in \mathbb{R}^{2}$.
Since the argmax operation is non-differentiable, no gradients are propagated
through this operator.
Accordingly, hard argmax landmarks are not used for training or physics-based
losses, but only for reporting final localization accuracy and visual evaluation.

\subsection{Ballistic Physics Model with Bounce Handling - Factor D and E}\label{sec:physics_model}
\label{fxy}
%%%%%%%%%%%%%%%%%%%%%%%%%%%%%%%%%%%%%%%%%%%%%%%%%%%%%%%%%%%%
This section describes the differentiable physics model used to regularize the landmarks coming from the differentiable bilinear expectation \ref{app:bilinear_expectation} outputs or using them for prediction of position, velocity, and bounce detection. The model operates directly in pixel space and is applied to short temporal windows
of three consecutive frames $(t\!-\!1, t, t\!+\!1)$.

\subsubsection{Dimensional Analysis: Physical to Frame Units}
\label{app:dim_analysis}

The data is generated using physical units ($g = 9.81\;\text{m/s}^2$, $S = 0.02\;\text{m/px}$, $\Delta t = 0.04\;\text{s/frame}$), but the physics module operates in \emph{frame units} to avoid numerical instability.
Converting gravity to pixel space gives $g_{\text{px}} = g / S = 490.5\;\text{px/s}^2$. The equivalent gravity in frame units is:
\begin{align}
g_{\text{frame}} = g_{\text{px}} \cdot \Delta t^2 = 490.5\;\text{px/s}^2 \times (0.04\;\text{s/frame})^2 = 0.7848\;\text{px/frame}^2
\end{align}
With $g_{\text{frame}}$ and $\Delta t = 1\;\text{frame}$, the per-frame displacement and velocity change are:
\begin{align}
\Delta y &= \tfrac{1}{2}\, g_{\text{frame}} \cdot 1^2 = 0.3924\;\text{px}, \qquad
\Delta v = g_{\text{frame}} \cdot 1 = 0.7848\;\text{px/frame}
\end{align}
This formulation avoids dividing by the small physical time step when computing velocity ($v = (p_1 - p_0)/\Delta t$), which would amplify position errors by a factor of $1/\Delta t = 25$.
Similarly, the maximum initial velocity $v_{\max} = 11.1\;\text{m/s}$ converts to $v_{\max,\text{frame}} = v_{\max} / S \cdot \Delta t = 22.2\;\text{px/frame}$.

\subsubsection{Assumptions}

The physics model assumes:
\begin{itemize}
    \item Frame-unit time step $\Delta t = 1$
    \item Known gravitational acceleration $g_{\text{frame}} = 0.7848\;\text{px/frame}^2$ (positive downward)
    \item No horizontal acceleration
    \item Elastic or inelastic wall collisions with coefficient of restitution $e$
    \item Known image boundaries and ball radius
\end{itemize}

\subsubsection{State Representation}

At each time step, the ball state is represented by position and velocity,
\[
\mathbf{p}(t) = (x(t), y(t)), \qquad
\mathbf{v}(t) = (v_x(t), v_y(t)).
\]

The physics module receives three estimated landmark positions (after scaling),
\[
\mathbf{p}_{t-1},\; \mathbf{p}_{t},\; \mathbf{p}_{t+1}.
\]

\subsubsection{Velocity Initialization}

An initial velocity is estimated using a forward difference from the left,
\[
\mathbf{v}_{t-1} =
\frac{\mathbf{p}_{t} - \mathbf{p}_{t-1}}{\Delta t}.
\]
This forward difference uses only the first two frames $(t\!-\!1, t)$ and avoids dependence on $p_{t+1}$, ensuring that collision detection in the subsequent Verlet step is not influenced by the third frame.

\subsubsection{Velocity-Verlet Integration with Bounce Detection}

The model advances the state using a forward Velocity--Verlet scheme.
The position update is given by
\[
\begin{aligned}
x^{+} &= x + v_x \Delta t, \\
y^{+} &= y + v_y \Delta t + \tfrac{1}{2} g \Delta t^2.
\end{aligned}
\]

A half-step vertical velocity is computed as
\[
v_y^{1/2} = v_y + \tfrac{1}{2} g \Delta t.
\]

Boundary collisions with the image walls and floor/ceiling are detected.
When a collision occurs, the corresponding velocity component is reflected and scaled,
\[
v \leftarrow -e\, v.
\]

The final vertical velocity update is
\[
v_y^{+} = v_y^{1/2} + \tfrac{1}{2} g \Delta t.
\]

A binary bounce indicator is recorded for each time step.
All conditional operations (collision detection, velocity reflection, boundary clamping) are implemented using \texttt{torch.where}, ensuring the physics module remains fully differentiable and gradients propagate through both bounce and non-bounce branches.

\subsubsection{Smooth-Trajectory Correction (No-Bounce Case)}

If no collisions are detected across the three-frame window, the trajectory is
assumed to be smooth.
In this case, the forward-integrated trajectory is replaced by an exact
constant-gravity parabola.

A second-order central velocity is computed as
\[
\mathbf{v}_{\mathrm{mid}} =
\frac{\mathbf{p}_{t+1} - \mathbf{p}_{t-1}}{2\Delta t}.
\]

The left vertical velocity is recovered as
\[
v_{y,t-1} = v_{\mathrm{mid},y} - g \Delta t.
\]

Exact positions are then reconstructed,
\[
\begin{aligned}
\mathbf{p}_{t} &=
\mathbf{p}_{t-1} + \mathbf{v}_{t-1} \Delta t
+ \tfrac{1}{2} g \Delta t^2, \\
\mathbf{p}_{t+1} &=
\mathbf{p}_{t-1} + \mathbf{v}_{t-1} (2\Delta t)
+ \tfrac{1}{2} g (2\Delta t)^2.
\end{aligned}
\]

Velocities are updated analytically,
\[
\mathbf{v}_{t} = \mathbf{v}_{t-1} + g \Delta t, \qquad
\mathbf{v}_{t+1} = \mathbf{v}_{t-1} + 2g \Delta t.
\]

This correction improves numerical accuracy and enforces exact physical consistency
when no impacts occur.

\subsubsection{Outputs}\label{physics-informed predictions}
The physics module returns these outputs (physics-informed predictions) in this order:
\begin{itemize}
    \item Predicted positions $\mathbf{p}_{\substack{t-1:t+1 \\ Bilinear}}$, $\mathbf{p}_{\substack{t-1:t+1 \\ arg-max}}$, $\mathbf{p}_{\substack{t-1:t+1 \\ Physics}}$
    \item Predicted velocities $\mathbf{v}_{t-1:t+1}$
    \item Bounce indicators $b_{t-1:t+1}$
\end{itemize}

All outputs are fully differentiable with respect to the input landmark coordinates (except the arg-max outputs $\mathbf{p}_{\substack{t-1:t+1 \\ arg-max}}$), enabling their use in the physics-informed loss functions described below.

\subsubsection{PILL loss --- Factor~D (Unsupervised physics)}\label{loss:pill}
The unsupervised physics-informed landmark loss (PILL) compares the physics-model prediction against the heatmap-derived landmark positions, encouraging physical consistency without requiring ground-truth trajectories.
The landmarks are forced to follow physically plausible (parabolic) trajectories, although the true trajectory is not known:
\begin{equation}
\mathcal{L}_{\mathrm{PILL}}
=
\frac{1}{BT}
\sum_{b,t}
\left\lVert
f\!\left(a\hat{\mathbf{c}}_{b,t}\right)
-
a\hat{\mathbf{c}}_{b,t}
\right\rVert_1,
\label{eq:loss_pill}
\end{equation}
where $f(\cdot)$ is the differentiable physics model and $a$ maps heatmap to image coordinates
($a\!=\!4$ for $56\!\to\!224$,
$a\!=\!2$ for $112\!\to\!224$,
$a\!=\!1$ at $224$).
The $\ell_1$ norm penalizes deviations between the predicted landmarks and their physics-refined counterparts.
This loss enforces self-consistency without using ground-truth annotations.

\subsubsection{PILLS loss --- Factor~E (Supervised physics)}\label{loss:pills}
The supervised variant (PILLS) extends PILL by comparing model predictions directly against ground-truth physical states --- positions, velocities, and bounce indicators:
\begin{equation}
\mathcal{L}_{\mathrm{PILLS}} =
\left\lVert \mathbf{p}
- \mathbf{p}^{\mathrm{gt}} \right\rVert_1
+
\left\lVert \mathbf{v}
- \mathbf{v}^{\mathrm{gt}} \right\rVert_1
+
0.01 \left\lVert b
- b^{\mathrm{gt}} \right\rVert_1.
\label{eq:loss_pills}
\end{equation}
Both PILL and PILLS are applied at all three spatial scales ($56$, $112$, $224$) and require the physics model to be fully differentiable.

\subsubsection{Factor~F (Noise condition)}
Factor~F controls the input noise level: $\sigma\!=\!0$ (clean) or $\sigma\!=\!1$ (noisy). This factor has no associated loss; it only affects input augmentation.

\subsubsection{Total loss}\label{loss:total}
\label{app:loss_details}
Physics losses are ramped via
$w(e) = w_{\min} +
(1 - w_{\min})\min(1, e/T)$.
PILL: $w_{\min}\!=\!0.01$, $T\!=\!10$.
PILLS: $w_{\min}\!=\!0.001$, $T\!=\!20$.
The total loss combines all components:
\begin{equation}
\mathcal{L}_{\mathrm{total}}(e)
=
\mathcal{L}_{\mathrm{AE}}
+
\mathcal{L}_{\mathrm{cone}}
+
\mathcal{L}_{\mathrm{hm}}
+
w_{\mathrm{PILL}}(e)\mathcal{L}_{\mathrm{PILL}}
+
w_{\mathrm{PILLS}}(e)\mathcal{L}_{\mathrm{PILLS}}.
\label{eq:loss_total}
\end{equation}

%%%%%%%%%%%%%%%%%%%%%%%%%%%%%%%%%%%%%%%%%%%%%%%%%%%%%%%%%%%%
\section{Full Factorial Tracking Results}\label{sec:full_factorial_results}
Tables~\ref{tab:l1-full-columns-f0} and~\ref{tab:l1-full-columns-f1} report the replicated factorial tracking errors across all $2^6{=}64$ configurations for all nine tracking outputs ($n=4$ replicates, mean $\pm$ std).

\subsection{Noise-free results ($F{=}0$)}

In the noise-free setting, configurations incorporating supervised physics-informed losses (PILLS, $E{=}1$) consistently achieve the overall lowest errors for the
bilinear interpolated predictions (B), the heatmap-based (H) and the physics-refined outputs (P) at
intermediate and high spatial resolutions (112 and 224). This indicates that explicit physical supervision
provides a strong inductive bias.

At the same time, several configurations without explicit landmark or physics-informed
supervision ($C{=}0$, $E{=}0$) already exhibit competitive performance in the absence of
noise, demonstrating that the underlying tracking task can be learned reliably from
clean visual input alone. Overall, the results suggest that while physics-informed supervision is not strictly
necessary in this regime, it leads to more consistent performance across architectural
choices, spatial resolutions, and output representations.

% Generated by: python src/scripts/d_generate_tables_aggregated.py
\begin{table*}[t]
\centering
\caption{Full factorial results ($\sigma=0$, 4 seeds, mean $\pm$ std).}
\label{tab:l1-full-columns-f0}
\resizebox{\textwidth}{!}{%
\begin{tabular}{clccccccccc}
\toprule
Row & Config & B$_{56}$ & B$_{112}$ & B$_{224}$ & H$_{56}$ & H$_{112}$ & H$_{224}$ & P$_{56}$ & P$_{112}$ & P$_{224}$ \\
\midrule
 0 & A0B0C0D0E0F0 & 7.7{\tiny$\pm$2.1} & ${>}50$ & ${>}50$ & 7.8{\tiny$\pm$2.1} & 9.2{\tiny$\pm$1.3} & 7.3{\tiny$\pm$2.2} & 6.9{\tiny$\pm$1.9} & ${>}50$ & ${>}50$ \\
 1 & A1B0C0D0E0F0 & ${>}20$ & ${>}20$ & ${>}20$ & ${>}20$ & ${>}20$ & ${>}10$ & ${>}20$ & ${>}20$ & ${>}20$ \\
 2 & A0B1C0D0E0F0 & ${>}50$ & ${>}20$ & ${>}20$ & ${>}50$ & ${>}50$ & ${>}20$ & ${>}20$ & ${>}20$ & ${>}20$ \\
 3 & A1B1C0D0E0F0 & ${>}50$ & ${>}20$ & ${>}50$ & ${>}50$ & ${>}50$ & ${>}20$ & ${>}50$ & ${>}20$ & ${>}50$ \\
 4 & A0B0C1D0E0F0 & 1.1{\tiny$\pm$0.0} & ${>}50$ & ${>}50$ & \underline{1.1{\tiny$\pm$0.0}} & ${>}10$ & ${>}10$ & 1.1{\tiny$\pm$0.0} & ${>}50$ & ${>}50$ \\
 5 & A1B0C1D0E0F0 & 2.0{\tiny$\pm$0.5} & 5.7{\tiny$\pm$3.5} & 4.2{\tiny$\pm$5.0} & 2.1{\tiny$\pm$0.5} & 1.2{\tiny$\pm$0.1} & 1.2{\tiny$\pm$0.1} & 2.0{\tiny$\pm$0.5} & 6.2{\tiny$\pm$4.0} & 4.3{\tiny$\pm$4.8} \\
 6 & A0B1C1D0E0F0 & 1.2{\tiny$\pm$0.1} & ${>}20$ & ${>}20$ & 1.2{\tiny$\pm$0.2} & ${>}20$ & ${>}20$ & 1.2{\tiny$\pm$0.1} & ${>}20$ & ${>}20$ \\
 7 & A1B1C1D0E0F0 & 1.4{\tiny$\pm$0.2} & 0.6{\tiny$\pm$0.0} & 0.4{\tiny$\pm$0.1} & 1.5{\tiny$\pm$0.2} & 1.0{\tiny$\pm$0.1} & \underline{0.8{\tiny$\pm$0.0}} & 1.5{\tiny$\pm$0.2} & 0.6{\tiny$\pm$0.0} & 0.5{\tiny$\pm$0.1} \\
 8 & A0B0C0D1E0F0 & ${>}10$ & ${>}50$ & ${>}50$ & ${>}10$ & 8.0{\tiny$\pm$1.9} & 8.8{\tiny$\pm$0.8} & ${>}10$ & ${>}50$ & ${>}50$ \\
 9 & A1B0C0D1E0F0 & ${>}10$ & ${>}50$ & ${>}50$ & ${>}10$ & ${>}10$ & ${>}10$ & ${>}10$ & ${>}50$ & ${>}50$ \\
10 & A0B1C0D1E0F0 & 9.3{\tiny$\pm$3.7} & ${>}20$ & ${>}20$ & 9.8{\tiny$\pm$4.2} & ${>}50$ & ${>}20$ & 9.2{\tiny$\pm$3.8} & ${>}20$ & ${>}20$ \\
11 & A1B1C0D1E0F0 & ${>}10$ & ${>}50$ & ${>}50$ & ${>}10$ & ${>}20$ & ${>}20$ & ${>}20$ & ${>}50$ & ${>}50$ \\
12 & A0B0C1D1E0F0 & \underline{1.1{\tiny$\pm$0.1}} & ${>}50$ & ${>}50$ & 1.2{\tiny$\pm$0.1} & 7.4{\tiny$\pm$1.8} & ${>}10$ & \underline{1.1{\tiny$\pm$0.1}} & ${>}50$ & ${>}50$ \\
13 & A1B0C1D1E0F0 & 1.4{\tiny$\pm$0.2} & ${>}20$ & ${>}20$ & 1.4{\tiny$\pm$0.2} & 1.4{\tiny$\pm$0.2} & 1.3{\tiny$\pm$0.2} & 1.4{\tiny$\pm$0.2} & ${>}20$ & ${>}20$ \\
14 & A0B1C1D1E0F0 & 1.1{\tiny$\pm$0.1} & ${>}10$ & ${>}20$ & 1.2{\tiny$\pm$0.1} & ${>}20$ & ${>}20$ & \underline{1.1{\tiny$\pm$0.1}} & ${>}10$ & ${>}20$ \\
15 & A1B1C1D1E0F0 & 1.7{\tiny$\pm$0.3} & ${>}10$ & ${>}10$ & 1.8{\tiny$\pm$0.3} & 1.1{\tiny$\pm$0.0} & 1.0{\tiny$\pm$0.1} & 1.7{\tiny$\pm$0.2} & ${>}10$ & ${>}10$ \\
16 & A0B0C0D0E1F0 & 8.4{\tiny$\pm$12.1} & ${>}50$ & ${>}50$ & 9.3{\tiny$\pm$12.4} & 9.0{\tiny$\pm$0.8} & 5.8{\tiny$\pm$1.6} & ${>}10$ & ${>}50$ & ${>}50$ \\
17 & A1B0C0D0E1F0 & 5.9{\tiny$\pm$3.8} & ${>}20$ & ${>}10$ & 6.4{\tiny$\pm$3.6} & ${>}50$ & ${>}10$ & 4.9{\tiny$\pm$4.6} & ${>}20$ & 7.8{\tiny$\pm$14.3} \\
18 & A0B1C0D0E1F0 & 1.8{\tiny$\pm$0.4} & ${>}20$ & ${>}20$ & 2.5{\tiny$\pm$0.5} & ${>}50$ & ${>}50$ & 1.7{\tiny$\pm$0.5} & ${>}20$ & ${>}20$ \\
19 & A1B1C0D0E1F0 & ${>}10$ & ${>}20$ & ${>}20$ & ${>}20$ & ${>}20$ & ${>}50$ & ${>}10$ & ${>}20$ & ${>}20$ \\
20 & A0B0C1D0E1F0 & 1.3{\tiny$\pm$0.1} & ${>}50$ & ${>}50$ & 1.7{\tiny$\pm$0.2} & 6.7{\tiny$\pm$2.3} & 7.2{\tiny$\pm$1.6} & 1.3{\tiny$\pm$0.1} & ${>}50$ & ${>}50$ \\
21 & A1B0C1D0E1F0 & 1.5{\tiny$\pm$0.3} & 2.1{\tiny$\pm$1.3} & 1.0{\tiny$\pm$0.9} & 1.9{\tiny$\pm$0.3} & 1.4{\tiny$\pm$0.1} & 1.4{\tiny$\pm$0.1} & 1.4{\tiny$\pm$0.2} & 0.6{\tiny$\pm$0.1} & 0.5{\tiny$\pm$0.1} \\
22 & A0B1C1D0E1F0 & 1.3{\tiny$\pm$0.1} & ${>}20$ & ${>}20$ & 1.8{\tiny$\pm$0.2} & ${>}50$ & ${>}20$ & 1.2{\tiny$\pm$0.1} & ${>}20$ & ${>}20$ \\
23 & A1B1C1D0E1F0 & 8.6{\tiny$\pm$14.9} & \underline{0.4{\tiny$\pm$0.0}} & \underline{0.3{\tiny$\pm$0.0}} & 9.2{\tiny$\pm$15.2} & 1.0{\tiny$\pm$0.0} & 1.0{\tiny$\pm$0.0} & ${>}10$ & 0.5{\tiny$\pm$0.0} & \underline{0.38} \\
24 & A0B0C0D1E1F0 & 5.0{\tiny$\pm$2.2} & ${>}50$ & ${>}50$ & 6.0{\tiny$\pm$2.3} & 6.9{\tiny$\pm$1.3} & 6.6{\tiny$\pm$1.1} & 4.8{\tiny$\pm$2.1} & ${>}50$ & ${>}50$ \\
25 & A1B0C0D1E1F0 & ${>}10$ & ${>}20$ & 0.9{\tiny$\pm$0.3} & ${>}20$ & ${>}20$ & 5.0{\tiny$\pm$2.3} & ${>}10$ & ${>}20$ & 0.8{\tiny$\pm$0.3} \\
26 & A0B1C0D1E1F0 & ${>}20$ & ${>}20$ & ${>}20$ & ${>}20$ & ${>}50$ & ${>}50$ & ${>}10$ & ${>}20$ & ${>}20$ \\
27 & A1B1C0D1E1F0 & ${>}10$ & ${>}50$ & 0.3{\tiny$\pm$0.0} & ${>}10$ & ${>}50$ & 2.4{\tiny$\pm$1.2} & ${>}10$ & ${>}50$ & \underline{0.4{\tiny$\pm$0.0}} \\
28 & A0B0C1D1E1F0 & 1.3{\tiny$\pm$0.1} & ${>}50$ & ${>}50$ & 1.8{\tiny$\pm$0.1} & 7.4{\tiny$\pm$1.0} & 6.9{\tiny$\pm$1.8} & 1.2{\tiny$\pm$0.1} & ${>}50$ & ${>}50$ \\
29 & A1B0C1D1E1F0 & 1.2{\tiny$\pm$0.2} & 0.7{\tiny$\pm$0.0} & 0.6 & 1.6{\tiny$\pm$0.3} & 1.5{\tiny$\pm$0.1} & 1.5{\tiny$\pm$0.1} & 1.2{\tiny$\pm$0.2} & 0.7{\tiny$\pm$0.0} & 0.6{\tiny$\pm$0.0} \\
30 & A0B1C1D1E1F0 & 1.2{\tiny$\pm$0.1} & ${>}20$ & ${>}20$ & 1.5{\tiny$\pm$0.1} & ${>}50$ & ${>}20$ & 1.1{\tiny$\pm$0.0} & ${>}20$ & ${>}20$ \\
31 & A1B1C1D1E1F0 & 1.6{\tiny$\pm$0.3} & \underline{0.4{\tiny$\pm$0.0}} & 0.3{\tiny$\pm$0.0} & 2.2{\tiny$\pm$0.4} & \underline{1.0{\tiny$\pm$0.1}} & 1.0{\tiny$\pm$0.0} & 1.5{\tiny$\pm$0.2} & \underline{0.42} & \underline{0.4{\tiny$\pm$0.0}} \\
\bottomrule
\end{tabular}
}% end resizebox
\end{table*}

\subsection{Noisy results ($F{=}1$)}

Under noisy conditions, overall tracking performance degrades across all configurations, but physics-informed supervision remains beneficial and not far away from the results obtained without noise. In this regime, configurations incorporating supervised physics-informed losses
(PILLS, $E{=}1$) consistently achieve the lowest errors, particularly for the
bilinear interpolated predictions (B) and the physics-refined outputs (P) at
intermediate and high spatial resolutions (112 and 224). Additionally, the models incorporating PILLS exhibit improved robustness compared to purely data-driven baselines, achieving lower errors across several outputs despite
the increased uncertainty introduced by noise.

% Generated by: python src/scripts/d_generate_tables_aggregated.py
\begin{table*}[t]
\centering
\caption{Full factorial results ($\sigma=1$, 4 seeds, mean $\pm$ std).}
\label{tab:l1-full-columns-f1}
\resizebox{\textwidth}{!}{%
\begin{tabular}{clccccccccc}
\toprule
Row & Config & B$_{56}$ & B$_{112}$ & B$_{224}$ & H$_{56}$ & H$_{112}$ & H$_{224}$ & P$_{56}$ & P$_{112}$ & P$_{224}$ \\
\midrule
32 & A0B0C0D0E0F1 & ${>}50$ & ${>}50$ & ${>}50$ & ${>}50$ & ${>}50$ & ${>}50$ & ${>}50$ & ${>}50$ & ${>}50$ \\
33 & A1B0C0D0E0F1 & ${>}50$ & ${>}50$ & ${>}50$ & ${>}50$ & ${>}50$ & ${>}50$ & ${>}50$ & ${>}50$ & ${>}50$ \\
34 & A0B1C0D0E0F1 & ${>}50$ & ${>}50$ & ${>}50$ & ${>}50$ & ${>}50$ & ${>}50$ & ${>}50$ & ${>}50$ & ${>}50$ \\
35 & A1B1C0D0E0F1 & ${>}50$ & ${>}50$ & ${>}50$ & ${>}50$ & ${>}50$ & ${>}50$ & ${>}50$ & ${>}50$ & ${>}50$ \\
36 & A0B0C1D0E0F1 & ${>}20$ & ${>}50$ & ${>}50$ & ${>}20$ & ${>}20$ & ${>}20$ & ${>}20$ & ${>}50$ & ${>}50$ \\
37 & A1B0C1D0E0F1 & ${>}50$ & ${>}50$ & ${>}20$ & ${>}50$ & ${>}20$ & ${>}20$ & ${>}50$ & ${>}50$ & ${>}20$ \\
38 & A0B1C1D0E0F1 & ${>}10$ & ${>}50$ & ${>}50$ & ${>}10$ & ${>}50$ & ${>}50$ & ${>}10$ & ${>}50$ & ${>}50$ \\
39 & A1B1C1D0E0F1 & ${>}20$ & 3.4{\tiny$\pm$1.9} & 2.2{\tiny$\pm$1.6} & ${>}20$ & \underline{1.0{\tiny$\pm$0.0}} & \underline{0.9{\tiny$\pm$0.1}} & ${>}20$ & 2.9{\tiny$\pm$1.3} & 2.0{\tiny$\pm$2.2} \\
40 & A0B0C0D1E0F1 & ${>}50$ & ${>}50$ & ${>}50$ & ${>}50$ & ${>}50$ & ${>}50$ & ${>}50$ & ${>}50$ & ${>}50$ \\
41 & A1B0C0D1E0F1 & ${>}50$ & ${>}50$ & ${>}50$ & ${>}50$ & ${>}50$ & ${>}50$ & ${>}50$ & ${>}50$ & ${>}50$ \\
42 & A0B1C0D1E0F1 & ${>}50$ & ${>}50$ & ${>}50$ & ${>}50$ & ${>}50$ & ${>}50$ & ${>}50$ & ${>}50$ & ${>}50$ \\
43 & A1B1C0D1E0F1 & ${>}50$ & ${>}50$ & ${>}50$ & ${>}50$ & ${>}50$ & ${>}50$ & ${>}50$ & ${>}50$ & ${>}50$ \\
44 & A0B0C1D1E0F1 & \underline{1.2{\tiny$\pm$0.0}} & ${>}50$ & ${>}50$ & \underline{1.3{\tiny$\pm$0.0}} & ${>}10$ & 10.0{\tiny$\pm$3.8} & \underline{1.2{\tiny$\pm$0.0}} & ${>}50$ & ${>}50$ \\
45 & A1B0C1D1E0F1 & ${>}50$ & ${>}50$ & ${>}50$ & ${>}50$ & 3.3{\tiny$\pm$1.1} & 3.0{\tiny$\pm$1.2} & ${>}50$ & ${>}50$ & ${>}50$ \\
46 & A0B1C1D1E0F1 & ${>}50$ & ${>}50$ & ${>}50$ & ${>}50$ & ${>}50$ & ${>}50$ & ${>}50$ & ${>}50$ & ${>}50$ \\
47 & A1B1C1D1E0F1 & 1.7{\tiny$\pm$0.7} & ${>}20$ & ${>}20$ & 1.9{\tiny$\pm$0.8} & 1.1{\tiny$\pm$0.1} & 1.0{\tiny$\pm$0.1} & 1.6{\tiny$\pm$0.6} & ${>}20$ & ${>}20$ \\
48 & A0B0C0D0E1F1 & ${>}50$ & ${>}50$ & ${>}50$ & ${>}50$ & ${>}50$ & ${>}50$ & ${>}50$ & ${>}50$ & ${>}50$ \\
49 & A1B0C0D0E1F1 & ${>}50$ & ${>}20$ & ${>}20$ & ${>}50$ & ${>}20$ & ${>}20$ & ${>}50$ & ${>}20$ & ${>}20$ \\
50 & A0B1C0D0E1F1 & ${>}50$ & ${>}50$ & ${>}50$ & ${>}50$ & ${>}50$ & ${>}50$ & ${>}50$ & ${>}50$ & ${>}50$ \\
51 & A1B1C0D0E1F1 & ${>}20$ & ${>}10$ & ${>}20$ & ${>}20$ & ${>}10$ & ${>}20$ & ${>}20$ & 4.1{\tiny$\pm$2.8} & ${>}20$ \\
52 & A0B0C1D0E1F1 & ${>}20$ & ${>}50$ & ${>}50$ & ${>}20$ & ${>}20$ & ${>}20$ & ${>}20$ & ${>}50$ & ${>}50$ \\
53 & A1B0C1D0E1F1 & ${>}50$ & ${>}20$ & ${>}20$ & ${>}50$ & ${>}20$ & ${>}20$ & ${>}50$ & ${>}20$ & ${>}20$ \\
54 & A0B1C1D0E1F1 & ${>}10$ & ${>}50$ & ${>}50$ & ${>}10$ & ${>}50$ & ${>}50$ & ${>}10$ & ${>}20$ & ${>}50$ \\
55 & A1B1C1D0E1F1 & ${>}10$ & \underline{0.4{\tiny$\pm$0.0}} & \underline{0.3{\tiny$\pm$0.0}} & ${>}10$ & 1.1{\tiny$\pm$0.1} & 1.1{\tiny$\pm$0.1} & ${>}10$ & \underline{0.5{\tiny$\pm$0.0}} & \underline{0.4{\tiny$\pm$0.0}} \\
56 & A0B0C0D1E1F1 & ${>}50$ & ${>}50$ & ${>}50$ & ${>}50$ & ${>}50$ & ${>}50$ & ${>}50$ & ${>}50$ & ${>}50$ \\
57 & A1B0C0D1E1F1 & ${>}50$ & ${>}20$ & ${>}20$ & ${>}50$ & ${>}20$ & ${>}20$ & ${>}50$ & ${>}20$ & ${>}20$ \\
58 & A0B1C0D1E1F1 & ${>}50$ & ${>}50$ & ${>}50$ & ${>}50$ & ${>}50$ & ${>}50$ & ${>}50$ & ${>}50$ & ${>}50$ \\
59 & A1B1C0D1E1F1 & ${>}50$ & ${>}10$ & ${>}20$ & ${>}50$ & ${>}10$ & ${>}20$ & ${>}50$ & ${>}10$ & ${>}20$ \\
60 & A0B0C1D1E1F1 & ${>}10$ & ${>}50$ & ${>}50$ & ${>}10$ & 7.3{\tiny$\pm$3.4} & ${>}10$ & ${>}10$ & ${>}50$ & ${>}50$ \\
61 & A1B0C1D1E1F1 & ${>}50$ & ${>}20$ & ${>}20$ & ${>}50$ & ${>}20$ & ${>}20$ & ${>}50$ & ${>}20$ & ${>}20$ \\
62 & A0B1C1D1E1F1 & ${>}20$ & ${>}20$ & ${>}50$ & ${>}20$ & ${>}50$ & ${>}50$ & ${>}20$ & ${>}20$ & ${>}50$ \\
63 & A1B1C1D1E1F1 & ${>}20$ & \underline{0.4{\tiny$\pm$0.0}} & 0.4{\tiny$\pm$0.0} & ${>}20$ & 1.1{\tiny$\pm$0.1} & 1.1{\tiny$\pm$0.1} & ${>}20$ & \underline{0.5{\tiny$\pm$0.0}} & \underline{0.4{\tiny$\pm$0.0}} \\
\bottomrule
\end{tabular}
}% end resizebox
\end{table*}

\subsection{Summary and discussion}

Taken together, the results across both noise regimes show that physics-informed
supervision improves robustness and consistency across resolutions and output
representations.

Despite some isolated low-error configurations, the overall results for the
unsupervised setting ($C{=}0$, $E{=}0$) are highly variable and do not exhibit
consistent performance across resolutions, outputs, or architectural choices.
This suggests that, in the absence of explicit landmark supervision or physics-based
constraints, the optimization problem is poorly conditioned, and performance becomes
sensitive to factors such as resolution, initialization, and output representation.

In the unsupervised setting, the linear ramping of the physics loss may delay the
enforcement of physical constraints, allowing visually plausible but physically
inconsistent solutions to form early in training, which helps explain the observed
performance variability.

While reconstruction quality is limited in this simplified experimental setup,
this aspect is not the primary focus of the study and remains closely connected to
the challenges of unsupervised and weakly supervised tracking.
At the same time, the results highlight that imposing physical structure at the
landmark and trajectory level provides a promising foundation for generative models,
where physical consistency is often more critical than pixel-level fidelity.
By constraining latent representations and predictions to obey known dynamics,
the proposed framework naturally supports generative settings in which trajectories
can be sampled or extrapolated in a manner that is both visually plausible and
physically consistent.
Further investigation is therefore warranted to explore tighter coupling between
representation learning, unsupervised tracking, and physics-constrained generative
modeling.

%These findings point toward several directions for improving unsupervised tracking.
%Stronger inductive structure—such as tighter temporal consistency terms, explicit
%trajectory regularization, curriculum-based resolution schedules, or hybrid losses
%that weakly encode physical constraints—may be necessary to stabilize learning.
%Additionally, more careful tuning of scale-dependent losses and normalization across
%outputs could reduce variance in the unsupervised regime.

\end{document}